\documentclass[]{fairmeta}

\usepackage{amsmath,amssymb}
\usepackage{adjustbox}
\usepackage{colortbl}
\usepackage{mdframed}
\usepackage{pifont}
\usepackage{tabularx}
\usepackage{xspace}

\newcommand{\our}{Flow2Act\xspace}
\newcommand{\cmark}{\textcolor{green}{\ding{51}}}
\newcommand{\xmark}{\textcolor{red}{\ding{55}}}

\allowdisplaybreaks[1]

\definecolor{green}{HTML}{009000}
\definecolor{red}{HTML}{EA4335}
\definecolor{oursrow}{HTML}{E8DFEE}     
\definecolor{experimentrow}{HTML}{F4F0F7}
\newcolumntype{R}[2]{%
    >{\adjustbox{angle=#1,lap=\width-(#2)}\bgroup}%
    l%
    <{\egroup}%
}
\newcommand{\rot}[1]{\multicolumn{1}{R{20}{3em}}{#1}}

\renewcommand{\paragraph}[1]{\vspace{1.25mm}\noindent\textbf{#1}}

\title{Flow2Act: Learning to Act through Geometrically Grounded Motion Anticipation}

\author[1,*]{Lijun Guo}
\author[2,3,*]{Haoyu Zhao}
\author[1, \ddagger]{Xianda Guo}
\author[4]{Chengwei Wu}
\author[2,3]{Yetong Zeng}
\author[5]{Chenglong Zhang}
\author[1]{\\Qian Zhou}
\author[2,3,\dagger]{Zhongyu Li}
\author[1,\dagger]{Hua Zou}

\affiliation[1]{Wuhan University,}
\affiliation[2]{Hong Kong Embodied AI Lab,}
\affiliation[3]{The Chinese University of Hong Kong,\\}
\affiliation[4]{Beihang University,}
\affiliation[5]{Nanjing University}

\contribution[*]{Equal contribution}
\contribution[\ddagger]{Project Lead}
\contribution[\dagger]{Co-corresponding authors}

\abstract{
Successful manipulation requires understanding scene geometry and anticipating how scene points will move during interaction.
Action imitation supervises robot commands, but does not explicitly shape the vision-language-action (VLA) model's representations of these spatial and temporal relationships.
We present \textbf{\our}, a framework that uses geometrically grounded motion anticipation to condition action generation.
Within a pretrained VLA, learnable future queries integrate current visual features with sparse observations from previous replanning steps. A frozen geometry teacher grounds the visual features via distillation, while paired 2D trajectories and relative-scale 3D displacement targets supervise the queries to anticipate where current scene anchors will move over the action horizon. The same query representations condition the native action generator through their layer-wise key-value states, providing a shared pathway for motion prediction and control.
At deployment, \our uses \textbf{only} RGB observations, robot states, and decision history, without teacher networks and dense prediction heads.
Evaluations across four pretrained VLA backbones and three simulation benchmarks show aggregate success gains over the corresponding base policies, complemented by experiments on six real-world manipulation tasks.
}

\renewcommand{\metadatalist}{%
    \centering
    \renewcommand{\arraystretch}{1.2}%
    \begin{tabular}{@{}cl@{}}
        \raisebox{-1.5pt}{\includegraphics[height=1.05em]{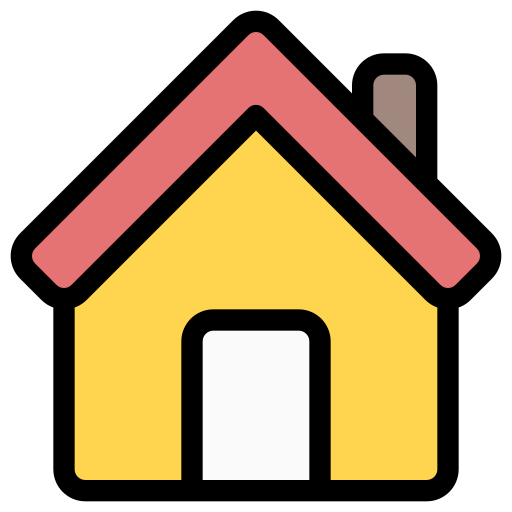}} &
            \url{https://shawnricardo.github.io/Flow2Act.github.io/} \\
        \raisebox{-1.5pt}{\includegraphics[height=1.05em]{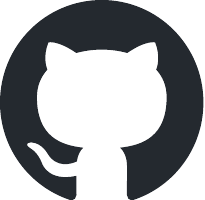}} &
            \url{https://github.com/ShawnRicardo/Flow2Act} \\
        \raisebox{-1.5pt}{\includegraphics[height=1.05em]{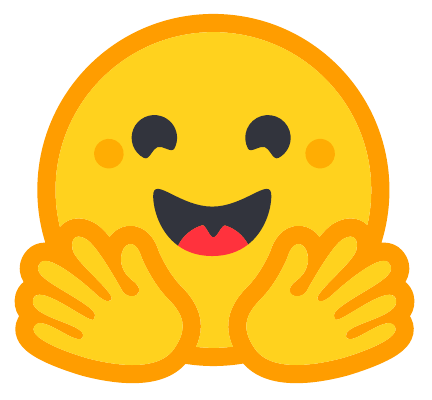}} &
            \url{https://huggingface.co/ShawnRicardo/Flow2Act-HF} \\
        \raisebox{-1pt}{\includegraphics[height=0.95em]{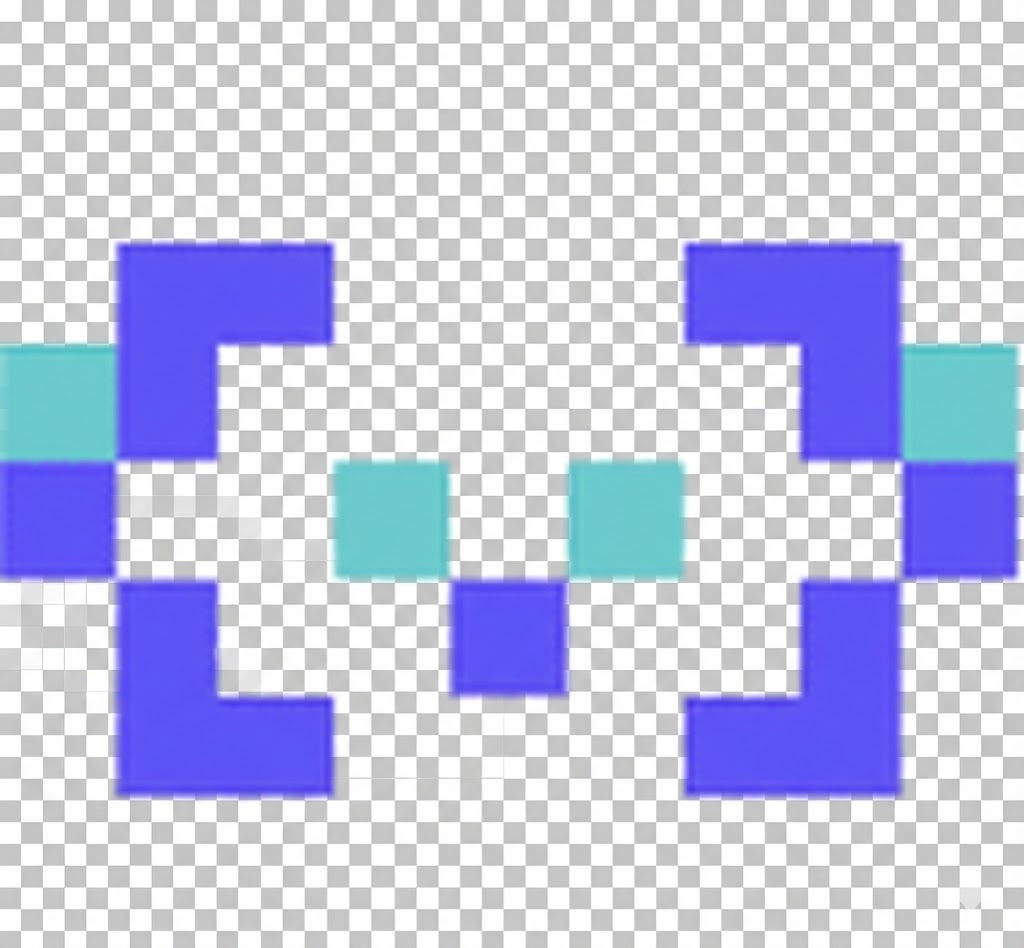}} &
            \url{https://modelscope.cn/models/ShawnRicardo/Flow2Act-MS}
    \end{tabular}%
}

\begin{document}
\thispagestyle{firstheader}
\maketitle
\pagestyle{empty}

\section{Introduction}
\label{sec:introduction}
Vision-language-action (VLA) models transfer semantic knowledge from large-scale vision-language pretraining to robot control~\citep{kim2024openvla,black2024pi_0,intelligence2025pi_,zhao2026rynnvla,brown2020language,hoffmann2022training,kaplan2020scaling,bjorck2025gr00t,cai2026xiaomi}.
Successful manipulation, however, requires more than recognizing objects and interpreting instructions: the robot must relate the scene's current geometry to how its contents will move during interaction~\citep{zhao2025physsplat,zhao2026high,zhao2026rynnworld,zhao2026rynnworldlatent}.
When closing a drawer, for example, similar gripper motions can leave the drawer stationary or move it along its rails, depending on the gripper's position relative to the drawer front and whether contact has been established.
Action imitation supervises the robot command, while leaving the geometry and scene motion only indirectly constrained.
This distinction matters when changes in object layouts, viewpoints, or robot initial states weaken appearance correlations learned from demonstrations.

% ====================================================================================
% ====================================================================================

\begin{table}[htbp]
    \centering
    % \small
    \setlength{\tabcolsep}{5pt}
    \begin{tabular}{ccccccc|l}
    \rot{RGB-only deployment}
    & \rot{Geometry-aware}
    & \rot{History-conditioned}
    & \rot{Future-state}
    & \rot{Point-level 3D}
    & \rot{Action-coupled}
    & \rot{Paired 2D/3D Motion} & \\
    \toprule

    % \cmark & \xmark & \xmark & \xmark & \xmark & \xmark & \xmark & RT-2~\citep{zitkovich2023rt} \\
    % \cmark & \xmark & \xmark & \xmark & \xmark & \xmark & \xmark & OpenVLA~\citep{kim2024openvla} \\
    \cmark & \xmark & \xmark & \xmark & \xmark & \xmark & \xmark & $\pi_{0.5}$~\citep{intelligence2025pi_} \\

    \xmark & \cmark & \xmark & \xmark & \xmark & \xmark & \xmark & PointVLA~\citep{li2026pointvla} \\
    \xmark & \cmark & \xmark & \cmark & \xmark & \xmark & \xmark & 3D-VLA~\citep{zhen20243d} \\

    \cmark & \xmark & \cmark & \xmark & \xmark & \xmark & \xmark & MemoryVLA~\citep{shi2026memoryvla} \\
    \cmark & \xmark & \xmark & \cmark & \xmark & \xmark & \xmark & GR-2~\citep{cheang2024gr} \\
    \cmark & \cmark & \cmark & \cmark & \cmark & \cmark & \xmark & FlowVLA~\citep{go2026flowvla} \\
    \cmark & \cmark & \cmark & \cmark & \xmark & \xmark & \xmark & InternW0-$\Delta$~\citep{miao2026internw0} \\

    \cmark & \cmark & \xmark & \xmark & \xmark & \xmark & \xmark & ROCKET~\citep{sun2026rocket} \\

    \midrule
    \rowcolor{experimentrow}
    \cmark & \cmark & \cmark & \cmark & \cmark & \cmark & \cmark & \textbf{\our} \\

    \bottomrule
    \vspace{2pt}
    \end{tabular}
    \caption{\textbf{Comparison with representative methods.}
        Columns indicate: RGB-only input at inference; geometry-aware
        representations; use of past decision-step observations;
        anticipation of future states; point-level 3D displacement
        supervision; predictive features that directly condition the
        action generator; and paired 2D trajectory and 3D displacement
        supervision.}
    \label{tab:capability_comparison}
\end{table}

% ====================================================================================
% ====================================================================================

Recent work explores complementary ways to enhance the spatial and temporal representations of VLA models, as summarized in Tab.~\ref{tab:capability_comparison}. Spatially grounded policies incorporate depth or point clouds~\citep{li2026pointvla,qu2025spatialvla}, or transfer geometric knowledge into visual features through training-time supervision~\citep{li2026spatial}.  
While these approaches improve spatial awareness without requiring explicit geometric inputs at deployment, static geometric supervision alone does not explicitly characterize how scene points evolve during physical interactions.
Predictive policies introduce temporal supervision through future images, videos, or latent representations~\citep{zhao2026rynnvla,shen2026videovla,miao2026internw0,wu2026foundation}, while memory-based policies use historical observations to inform action generation~\citep{shi2026memoryvla,shi2026memoryvla++}. More directly, Pri4R~\citep{kim2026pri4r} and FlowVLA~\citep{go2026flowvla} supervise future 3D point motion, demonstrating the effectiveness of explicit motion targets for policy learning.
Such supervision ties future motion to scene points and provides spatial and temporal cues for action generation.
These advances motivate combining complementary evidence for motion anticipation: current geometry provides spatial context, while recent history reveals changes that a single observation cannot resolve. We investigate how to integrate this evidence into predictive representations for action generation:

\vspace{5pt}
\begin{mdframed}[backgroundcolor=oursrow!100, linewidth=0.9pt, linecolor=oursrow, roundcorner=10pt]
\textit{How can a pretrained VLA combine current geometry and recent interaction history into future-motion representations that directly condition action generation, without explicit 3D inputs at deployment?}
\end{mdframed}
\vspace{5pt}

% Recent work provides two complementary starting points.
% Spatially grounded policies incorporate depth or point clouds~\citep{li2026pointvla,qu2025spatialvla}, or transfer geometric knowledge into image features through training-time supervision~\citep{li2026spatial}.
% The latter approach is particularly attractive because the geometry teacher can be discarded at deployment.
% Yet learning where scene points are does not by itself supervise where they will move~\citep{kim2026pri4r,go2026flowvla}.
% Predictive policies address future evolution through image, video, or latent prediction~\citep{zhao2026rynnvla,shen2026videovla}.
% MemoryVLA~\citep{shi2026memoryvla} and MemoryVLA++~\citep{shi2026memoryvla++} combines retrieved history with imagined features for action generation~\citep{wang2026lamp,xiong2026geomvla,xiao2026ava}, while InternW0-$\Delta$~\citep{miao2026internw0} learns video-latent changes and distills geometry and motion descriptors~\citep{cheang2024gr,shen2026videovla}.
% LingBot-VLA 2.0~\citep{wu2026foundation} likewise uses current and future feature targets to encourage predictive understanding.
% These advances motivate a more specific prediction target: a low frame-reconstruction or feature-alignment loss does not directly specify whether a particular scene point reaches the correct location at each future step. We therefore ask:

Our hypothesis is that representations trained to anticipate scene-point motion can provide useful conditioning for action generation. Anchoring predictions to current scene points gives the supervision an explicit spatial and temporal reference: \textbf{which points move}, and \textbf{when and where they move} over the action horizon. 
We use the same representations for motion prediction and action conditioning, so that the action generator can draw on features explicitly trained to anticipate scene changes.

We introduce \textbf{\our}, a framework that implements this predictive interface through learnable future queries within pretrained VLA policies. Each query corresponds to a future offset in the action horizon and integrates current visual features with sparse historical context.
% Static geometry distillation aligns the VLM's visual representations with hierarchical features from a frozen 3D teacher~\citep{lin2025depth}, while a sparse historical memory supplies past observations and robot states at the policy's replanning cadence. Paired motion supervision trains the query representations to predict the displacements of current-image anchors.
We train the future-query representations to predict the motion of current-image anchors over the action horizon. To ground this motion anticipation in scene geometry, static geometry distillation aligns the VLM's visual representations with hierarchical features from a frozen 3D teacher~\citep{lin2025depth}. A sparse historical memory further supplies past visual features and robot states sampled at the policy's replanning cadence.
Offline Track4World~\citep{lu2026track4world} provide 2D trajectory targets, which are lifted using the teacher-estimated geometry to obtain paired, relative-scale 3D displacement targets. The same query representations condition the native action generator via their layer-wise key-value states. Future observations are used only to construct training targets. At deployment, the future queries and action conditioning remain active, while teacher networks and dense prediction heads are removed. 
Our contributions are as follows:
\begin{itemize}
    \item We introduce \textbf{\our}, a framework in which future-query representations within a pretrained VLA receive geometrically grounded scene-motion supervision and directly condition the action generator.
    \item We develop paired 2D and relative-scale 3D displacement supervision for current-image anchors across the action horizon, where motion anticipation is informed by static geometry distillation and sparse decision history.
    \item We evaluate the framework across \textbf{4} pretrained VLA backbones on \textbf{3} simulation benchmarks, together with \textbf{6} real-world manipulation tasks, demonstrating aggregate improvements over the corresponding base policies.
\end{itemize}

\section{Related Work}
\label{sec:related_work}

\subsection{3D Geometry for VLA}
VLA policies~\citep{kim2024openvla,black2024pi_0,intelligence2025pi_,zhao2026rynnvla,bjorck2025gr00t,cai2026xiaomi} benefit from vision-language pretraining, but precise manipulation also depends on spatial relations that are not explicit in ordinary image features. 
A first family supplies geometry as an additional input at deployment time~\citep{qu2025spatialvla,zhen20243d}.
PointVLA~\citep{li2026pointvla} injects point-cloud features into a pretrained action model, while BridgeVLA~\citep{li2026bridgevla} projects point clouds into multiple image views and predicts actions as 2D heatmaps in the projected views. These designs differ in how geometric information is obtained and presented to the policy at inference.

A separate line of work uses geometry as supervision rather than an additional deployment input~\citep{zhao2026gem}. Spatial Forcing~\citep{li2026spatial} aligns intermediate VLA features with a pretrained 3D model~\citep{wang2025vggt}, and ROCKET~\citep{sun2026rocket} extends this idea to multiple layers using a shared projector. 
Our geometry objective likewise distills from a frozen teacher~\citep{lin2025depth}, but uses a single adapter shared across layers.
Geometry supervision constrains \emph{where} points are, while the motion objectives constrain \emph{how they move}.

\subsection{Future Motion and Temporal Context}
VLA models pair a VLM backbone with a lightweight action expert that decodes action chunks from the backbone's hidden states~\citep{black2024pi_0,bjorck2025gr00t,intelligence2025pi_}.
The question is what additional predictive signal, if any, should condition this expert beyond the current observation.
One line predicts future appearance or video latents~\citep{liao2026genie,zhang2026foreact}. For instance, GR-2~\citep{cheang2024gr} and VideoVLA~\citep{shen2026videovla} couple video prediction with action learning. 

A more compact alternative predicts motion directly: optical flow or 2D point trajectories~\citep{zhong2026flowvla,chen2026flowwam,li2026fomovla,bharadhwaj2024track2act,zheng2025tracevla}, or 3D point tracks~\citep{qian2026geopredict,wang2026lamp,xiong2026geomvla,gao2026focal}.
However, these methods typically require a separate motion expert or decoder~\citep{wang2026lamp,xiong2026geomvla}, match query representations to teacher features rather than supervising explicit outputs~\citep{wang2026track4action,miao2026internw0}, or rely on temporally dense tracks expressed in a world-centric frame~\citep{kim2026pri4r,go2026flowvla}.
Our future queries avoid all three: they reside in the VLM's causal sequence, are supervised by paired 2D and relative-scale 3D displacements of current-image anchors, and condition the native action expert through layer-wise key--value states without additional modules or teachers at inference.

History provides information that a single observation cannot reveal~\citep{lin2026hif,ding2026temporal}. MemoryVLA~\citep{shi2026memoryvla} and MemoryVLA++~\citep{shi2026memoryvla++} retrieve past features from a memory bank, while Temporal Forcing~\citep{ding2026temporal} compresses history into tokens and aligns them with causal 4D features to supervise what the history should \emph{retain}.
We instead use history to inform what the future queries \emph{predict}: sparse observations from prior replanning decisions are placed directly in the VLM sequence, and future queries attend to them through native causal self-attention in a single forward pass.

\section{Method}
\label{sec:method}

\begin{figure}[htbp]
\centering
\includegraphics[width=\linewidth]{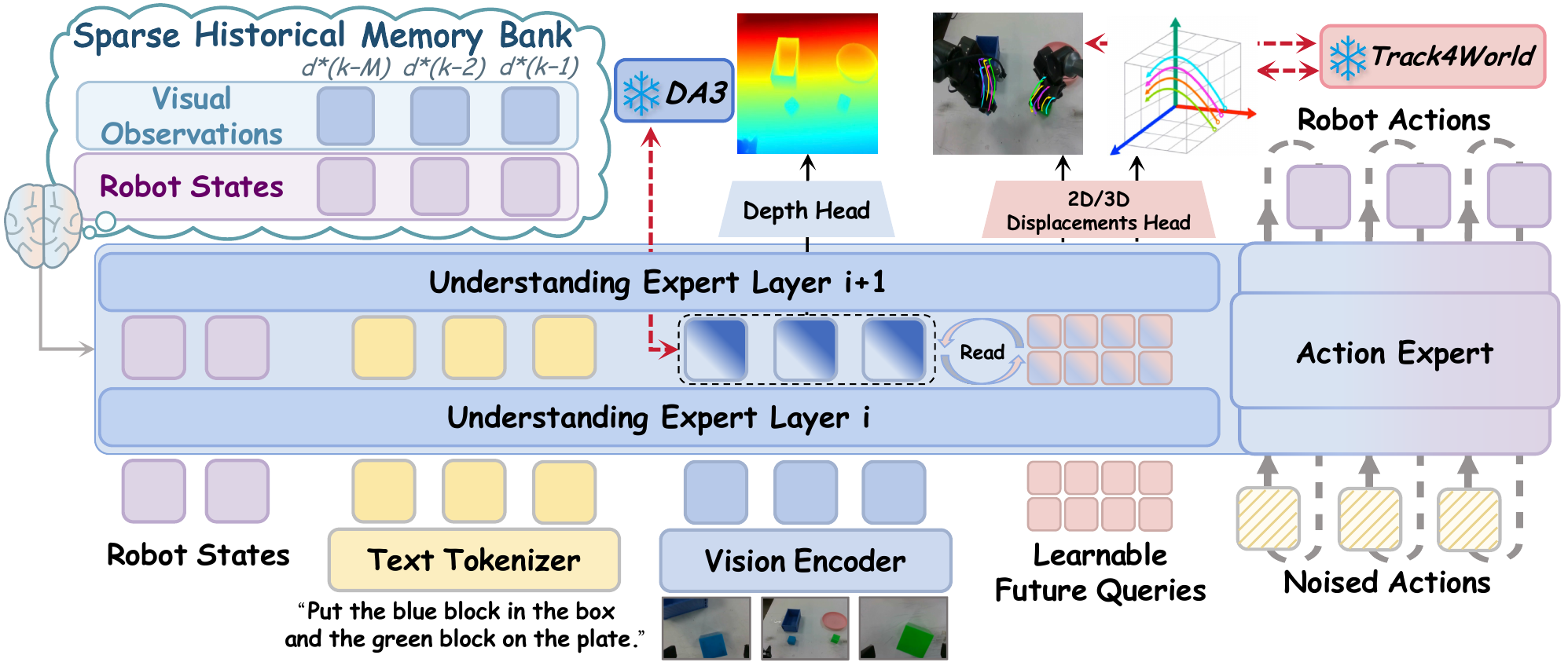}
\vspace{2pt}
\caption{\textbf{Overview of \our.} The VLM integrates current visual-language observations with sparse historical observations and robot states through learnable future queries.
DA3 and Track4World provide geometric and motion supervision during training, respectively.
The resulting future representations predict current-anchor 2D/3D displacements and condition the Action Expert for robot action generation.
Teacher models and dense motion prediction heads are removed at inference.
\textbf{\textcolor[HTML]{C81D31}{Red arrows}} denote training-time distillation from DA3 (geometry) and Track4World (motion).}
\label{fig:method-pipeline}
\end{figure}

As illustrated in Fig.~\ref{fig:method-pipeline}, \our couples motion anticipation with action generation through learnable future queries in a shared VLM.
At control step $t$, the queries combine visual features shaped by geometry distillation with sparse historical memory to predict future 2D and 3D point displacements.
The resulting query representations condition the native action model.

\subsection{Predicting Future Motion within the VLM}
\label{sec:future}

% \noindent\textbf{Future queries.}
% For an $N$-step action horizon, we introduce learnable queries $Q=[q_1;\ldots;q_N]\in\mathbb R^{N\times d}$, where $d$ is the VLM hidden width and query $h$ corresponds to time $t+h$.
% Given context $X_t$ comprising the current image--language prefix, robot state, and available decision history, we obtain
% \begin{equation}
%  F_t=\operatorname{VLM}_{Q}\!\left([X_t;Q]\right),
%  \label{eq:queries}
% \end{equation}
% where $\operatorname{VLM}_{Q}$ extracts the final normalized hidden states at the query positions, yielding $F_t=[f_{t,1};\ldots;f_{t,N}]$.

% At layer $b$, the hidden state at query position $h$ is projected to an attention query $\mathbf q_{t,h}^{(b)}$.
% Under native causal attention, the corresponding keys $K_{t,\leq h}^{(b)}$ and values $V_{t,\leq h}^{(b)}$ are projected from the hidden states of $X_t$ and query positions $1,\ldots,h$.
% For one attention head,
% \begin{equation}
%  c_{t,h}^{(b)}
%  =
%  \operatorname{softmax}\!\left(
%  \frac{
%  \mathbf q_{t,h}^{(b)}
%  \bigl(K_{t,\leq h}^{(b)}\bigr)^\top
%  }{\sqrt{d_{\mathrm{head}}}}
%  \right)
%  V_{t,\leq h}^{(b)},
%  \label{eq:query-context-attention}
% \end{equation}
% where $d_{\mathrm{head}}$ is the head width.
% All query representations are computed in a single VLM forward pass.
\noindent\textbf{Future queries.}
For an $N$-step action horizon, we introduce learnable queries $Q=[q_1;\ldots;q_N]\in\mathbb R^{N\times d}$, where $d$ is the VLM hidden width and query $h$ corresponds to time $t+h$. Given context $X_t$ comprising the current image--language prefix, robot state, and available decision history, we obtain
\begin{equation}
 F_t=\operatorname{VLM}_{Q}\!\left([X_t;Q]\right),
 \label{eq:queries}
\end{equation}
where $\operatorname{VLM}_{Q}$ extracts the final normalized hidden states at the query positions, yielding $F_t=[f_{t,1};\ldots;f_{t,N}]$. Under native causal attention, query $h$ attends to $X_t$ and query positions $1,\ldots,h$. All query representations are computed in a single VLM forward pass.

\noindent\textbf{Motion readout.}
For a current main-view anchor $p=(x,y)$ in image-pixel coordinates, let $g_t(p)$ be the final main-view VLM feature after channel projection and bilinear resampling at $p$, and $e(p)$ its spatial encoding. We combine these features with the future-query representation
through a projection $W_F$ that matches the visual feature width:
\begin{equation}
 \begin{aligned}
 u_{t,h}(p)
 &=\mathcal R\!\left(g_t(p)+f_{t,h}W_F+e(p)\right),\\
 \widehat d_{t,h}(p)
 &=\mathcal H_{\mathrm{2D}}\!\left(u_{t,h}(p)\right),\\
 \widehat\delta_{t,h}(p)
 &=\mathcal H_{\mathrm{3D}}\!\left(u_{t,h}(p)\right).
 \end{aligned}
 \label{eq:motion-errors}
\end{equation}
The readout $\mathcal R$ consists of LayerNorm, a linear layer, and GELU.
Two linear heads predict normalized 2D image-plane displacements and relative-scale 3D displacements, respectively. The readout and both heads are shared across anchors and future offsets.

\noindent\textbf{Paired motion supervision.}
We supervise both readouts using paired targets constructed from Track4World~\citep{lu2026track4world} correspondences.
Let $q^*_{t,h}(p)$ denote the correspondence of anchor $p$ at time $t+h$.
For image width $W$ and height $H$, the normalized 2D displacement target is
\begin{equation}
 d^*_{t,h}(p)
 =\operatorname{diag}(W^{-1},H^{-1})
 \bigl(q^*_{t,h}(p)-p\bigr).
 \label{eq:targets}
\end{equation}
% All future offsets use the same current anchor, so the targets represent displacements from the current observation rather than between consecutive future frames.

For 3D supervision, let $P_t(p)$ and $P_{t+h\rightarrow t}(q)$ denote the current and transformed future points, respectively.
The relative-scale 3D displacement target is
\begin{equation}
 \delta^*_{t,h}(p)
 =
 \frac{
 P_{t+h\rightarrow t}\!\left(q^*_{t,h}(p)\right)-P_t(p)
 }{r_t},
 \label{eq:motion3d}
\end{equation}
where $r_t$ is the median valid depth of the current main view obtained from the same Track4World reconstruction. This normalization expresses displacement relative to the
current scene depth. The paired 2D and 3D displacements targets correspond to the same anchor and future offset, with all displacements measured from the current observation rather than between consecutive future frames.

\noindent\textbf{Training objectives.}
We train the two readouts using
\begin{align}
 \mathcal L_{\mathrm{2D}}
 &=
 \left\langle
 \operatorname{SmoothL1}\!\left(\widehat d,d^*\right)
 \right\rangle_{\mathrm{balanced}},
 \label{eq:motion2d-loss}\\
 \mathcal L_{\mathrm{3D}}
 &=
 \left\langle
 \operatorname{SmoothL1}\!\left(\widehat\delta,\delta^*\right)
 \right\rangle_{\mathrm{valid}}.
 \label{eq:motion3d-loss}
\end{align}
The 2D loss balances the mean losses of moving and low-motion anchors, while the 3D loss averages over valid targets.
Validity is assessed independently for each target type and future offset, preserving valid 2D supervision when the corresponding 3D target is invalid.

%%%%%%%%%%% Static 3D %%%%%%%%%%%%%%%

\subsection{Grounding the VLM in 3D Geometry}
\label{sec:static}
The motion objectives in Sec.~\ref{sec:future} supervise future point displacements through motion readouts, but do not directly align intermediate VLM visual features with geometric targets.
To provide a complementary constraint on current scene structure, we introduce static geometry distillation, which aligns the VLM's visual representations with features from a pretrained 3D foundation model~\citep{wang2025vggt,lin2025depth}.

Specifically, we use frozen DA3~\citep{lin2025depth} as the geometry teacher.
During policy training, DA3 jointly encodes the current camera views to produce hierarchical geometric features.
For each selected student--teacher layer pair, an adapter $\mathcal A$, shared across layers and camera views, maps student visual tokens into the teacher feature space.
The corresponding teacher features are downsampled to the student token grid for spatial alignment within each view.
We then minimize the cosine distance between the projected student features and their aligned teacher targets:
\begin{equation}
 \mathcal L_{\mathrm{geo}}=\operatorname{mean}\!\left[
 1-\cos\!\left(
 \mathcal A(z^{\mathrm{VLM}}),
 \widetilde z^{\mathrm{DA3}}
 \right)
 \right],
 \label{eq:static-error}
\end{equation}
where $z^{\mathrm{VLM}}$ denotes a student visual token and $\widetilde z^{\mathrm{DA3}}$ its spatially aligned teacher target from the paired layer.
This feature-level objective complements displacement supervision by explicitly grounding the visual representations used for motion anticipation.
Future queries attend to these student representations through the VLM's native self-attention, allowing the distilled geometric information to inform motion anticipation.
The adapter is used solely to compute $\mathcal L_{\mathrm{geo}}$; neither its outputs nor the teacher features are supplied as inputs to the future queries or motion readouts.

\subsection{Enhancing Motion Learning with Historical Context}
\label{sec:memory}

A single observation may not fully reveal ongoing motion or recent interaction changes.
We therefore equip future queries with sparse historical context, providing temporal evidence for predicting the future motion of current-image anchors.

Let $o_\tau=(I_\tau^{1:V},s_\tau)$ denote the multi-view RGB observations and robot state at control step $\tau$, and let $d_k$ denote the control step of the $k$-th replanning decision. At $t=d_k$, we maintain a first-in, first-out (FIFO) buffer of the $M$ preceding decision observations:
\begin{equation}
 \mathcal M_k=
 \left(o_{d_{k-1}},\ldots,o_{d_{k-M}}\right).
 \label{eq:memory}
\end{equation}
For complete execution windows of length $E$, these observations occur at $t-E,\ldots,t-ME$.
During training, we sample history from the same demonstration episode at the deployment decision interval, masking unavailable observations. At inference, the buffer is updated only at replanning decisions and cleared at episode boundaries or resets.

\noindent\textbf{Historical token construction.}
Historical images are re-encoded using the shared VLM, with gradients stopped through this feature-extraction pass.
The extracted visual tokens are mapped into the VLM token space using a trainable memory projection, augmented with spatial, camera, and temporal encodings, and combined with the corresponding historical robot-state tokens.
For historical observation $j$, the visual memory tokens are:
\begin{equation}
 m_{j,c,p}=
 W_m\operatorname{LN}(F_{j,c,p})+b_m
 +e_p+e_c+e_j,
 \label{eq:memory-tokens}
\end{equation}
where $F_{j,c,p}$ denotes the extracted visual feature at spatial position $p$ from camera $c$, $W_m$ and $b_m$ are the shared projection parameters, and $e_p$, $e_c$, and $e_j$ encode spatial position, camera identity, and observation age, respectively.
These visual and robot-state tokens form the historical context $B_t$.
The memory projection and shared VLM remain trainable through the main forward pass.

\noindent\textbf{Integration with future queries.}
Let $X_t^{\mathrm{cur}}$ denote the native current image--language prefix and $e_s(s_t)$ the current-state projection.
The context in Eq.~\eqref{eq:queries} is:
\begin{equation}
 X_t=
 \left[X_t^{\mathrm{cur}};B_t;e_s(s_t)\right].
 \label{eq:context}
\end{equation}
Future queries attend to the current prefix, historical context, and current robot state through the VLM's native causal self-attention.
Current-prefix representations do not directly attend to historical context in this forward pass.
The history-conditioned queries retain the same motion-readout and action-conditioning interfaces, allowing historical context to inform both future motion prediction and action generation.

% ======================================================================
% ======================================================================
\begin{table}[t]
\centering
\small
\renewcommand{\arraystretch}{1.2}
\setlength{\tabcolsep}{12pt}
\vspace{2pt}
\caption{Benchmark policy settings. $N$ and $E$ denote the action and execution horizons, respectively.}
\label{tab:app-backbone-config}
\begin{tabular}{@{}l l c c l@{}}
\toprule
\textbf{Backbone} & \textbf{Benchmarks} & $N$ & $E$ & \textbf{Action formulation} \\
\midrule
GR00T-N1.7       & LIBERO, Plus           & 16 & 8  & Discrete tokens \\
$\pi_{0.5}$      & LIBERO, Plus, RoboTwin & 50 & 10 & Flow-matching continuous \\
Xiaomi-Robotics-0 & LIBERO, Plus          & 50 & 10  & Continuous delta-EEF \\
LingBot-VLA 2.0  & RoboTwin               & 25 & 5  & Continuous delta-EEF \\
\bottomrule
\end{tabular}
\end{table}

% ======================================================================
% ======================================================================

\subsection{Joint Training and Inference}
\label{sec:joint}
The native action model is conditioned on the layer-wise VLM key--value states of the current image--language prefix and future queries.
We jointly optimize the native action loss with geometric and motion supervision:
\begin{equation}
 \begin{aligned}
 \mathcal L={}&\mathcal L_{\mathrm{action}}
 +\lambda_{\mathrm{geo}}\mathcal L_{\mathrm{geo}}
 +\lambda_{\mathrm{2D}}\mathcal L_{\mathrm{2D}}
 +\lambda_{\mathrm{3D}}\mathcal L_{\mathrm{3D}}.
 \end{aligned}
 \label{eq:joint}
\end{equation}
Teacher outputs and future observations are training-only.

\section{Experiments}
\label{sec:exp-main}
\subsection{Implementation Details}
\noindent\textbf{Training.}
We initialize each backbone from its publicly released pretrained checkpoint. For each training benchmark, we jointly fine-tune a single policy across all tasks, retaining the backbone's native action formulation and generator.
The training objective combines the native action loss with paired 2D/3D motion supervision derived from Track4World trajectories~\citep{lu2026track4world} and static geometry distillation from a frozen DA3 teacher~\citep{lin2025depth}.
Motion prediction spans the same $N$-step horizon as action generation, and historical observations are sampled at the deployment replanning cadence determined by the execution
horizon $E$.
Training hyperparameters and evaluation protocols are provided in Secs.~\ref{app:hyperparams} and~\ref{app:eval-protocol}, respectively.

\noindent\textbf{Evaluation protocol.}
At each decision step the policy predicts $N$ actions and executes up to the first $E$ before replanning. Inference uses only the student policy and its decision history; teacher networks and dense auxiliary readouts are absent. The history bank is cleared at every episode boundary. Further evaluation details (episode counts, seeding, metric definitions) are in Appendix.

\noindent\textbf{Benchmark-specific training configuration.}
For LIBERO / LIBERO-Plus, we use the four standard ten-task suites (Spatial, Object, Goal, Long) and jointly train a single policy across all forty tasks. Each episode provides a main-camera and a wrist-camera RGB image, a natural-language instruction, and the robot proprioceptive state. LIBERO-Plus is used exclusively for evaluation: policies trained on LIBERO are tested zero-shot under seven perturbation dimensions (camera viewpoint, robot initial state, language paraphrase, lighting, background texture, sensor noise, object layout).

For RoboTwin 2.0, we jointly fine-tune a single policy per backbone on 2{,}500 Clean and 25{,}000 Randomized bimanual demonstrations. Observations consist of stereo main-camera and wrist-camera RGB, language, and dual-arm proprioceptive state. Evaluation is reported separately under Clean and Randomized conditions.

\noindent\textbf{Backbone-specific settings.}
Tab.~\ref{tab:app-backbone-config} lists per-backbone choices. The native action formulation and generator are inherited from the pretrained checkpoint. Future queries, historical context, and auxiliary supervision are added as described in Sec.~\ref{sec:method}.

For \our variants, the auxiliary supervision is configured identically across backbones: four statically aligned layers, a shared convolutional adapter, both 2D trajectory and 3D displacement motion objectives, and an observation-context length of $L=3$ (current plus two past decision observations).

\subsection{Simulation Benchmarks}
\label{sec:exp-benchmarks}

\noindent\textbf{LIBERO and LIBERO-Plus.}
LIBERO~\citep{liu2023libero} evaluates language-conditioned single-arm manipulation across spatial, object, goal, and long-horizon task suites. LIBERO-Plus~\citep{fei2025libero} uses the same tasks but evaluates robustness under perturbations such as viewpoint, robot state, language, lighting, background, layout, and sensor noise. We use LIBERO to verify basic manipulation competence and LIBERO-Plus to measure robustness beyond in-distribution execution.

\noindent\textbf{RoboTwin 2.0.}
RoboTwin 2.0~\citep{chen2025robotwin} provides bimanual manipulation tasks with domain randomization. The comparison reports Clean and Randomized success rates separately, together with their mean.

\subsection{Real-World Robot Tasks}
\label{sec:exp-real-tasks}
\label{sec:real_robot_setup}
\label{app:real-world-detailed}

\noindent\textbf{Real-world robot setup.}
We evaluate on a PiperX dual-arm system using one head-mounted and two wrist-mounted RGB cameras. Hardware details and the platform illustration are provided in Appendix~\ref{app:real-world-setup}.

\noindent\textbf{Training data.}
We collect 500 demonstrations for each of the six task categories,
for 3{,}000 demonstrations in total. Real-world deployment uses
the $\pi_{0.5}$ backbone with its native action generator.

\noindent\textbf{Tasks.}
We deploy \our with the $\pi_{0.5}$ backbone (\our-$\pi_{0.5}$) on six real-world manipulation tasks.
\begin{itemize}
\item \textbf{Fruit Placement} requires selecting the strawberry or banana specified by the language instruction and placing it on a plate.
\item \textbf{Block-to-Cup} involves placing two differently colored blocks into a cup in the instructed order.
\item \textbf{Stack Blocks} requires stacking four blocks into the configuration specified by the language instruction.
\item \textbf{Object Transfer} requires both arms to simultaneously place two blocks, each onto a plate or into a box.
\item \textbf{Drawer Storage} involves opening a drawer, placing an orange block inside, and closing it.
\item \textbf{Shirt Folding} requires coordinated bimanual folding of a shirt.
\end{itemize}
Representative task executions are shown in Fig.~\ref{fig:exp-real-world} in Appendix~\ref{app:real-world-executions}.
\noindent\textbf{Evaluation.}
We conduct 20 trials per method per task category, not separately
for each language configuration. A trial succeeds only when the
entire instruction is completed correctly, including object selection,
placement targets, execution order, and all required manipulation
stages. Both arms must complete their assigned operations in
bimanual tasks; partial completion is counted as failure.

\begin{table}[t]
\centering
\small
\renewcommand{\arraystretch}{1.12}
\setlength{\tabcolsep}{3pt}
\caption{LIBERO~\citep{liu2023libero} success rate ($\%$, $\uparrow$).
Our variants jointly train one policy on all four suites.
Avg. is the suite mean rounded to the displayed precision.}
\vspace{2pt}
\label{tab:exp-libero}
\begin{tabularx}{\textwidth}{
    >{\raggedright\arraybackslash}p{0.34\textwidth}
    *{5}{>{\centering\arraybackslash}X}
}
\toprule
\textbf{Model}
& \textbf{Spatial}
& \textbf{Object}
& \textbf{Goal}
& \textbf{Long}
& \textbf{Avg.} \\
\midrule
X-VLA~\citep{zheng2026x}
& 98.20 & 98.60 & 97.80 & 97.60 & 98.1 \\
ABot-M0~\citep{yang2026abot}
& 98.80 & 99.80 & 99.00 & 96.60 & 98.6 \\
StarVLA-OFT~\citep{community2026starvla}
& 97.80 & 98.60 & 96.20 & 93.80 & 96.60 \\
Galaxea G0.5~\citep{liu2026g0}
& 98.40 & 100.00 & 98.60 & 98.60 & 98.90 \\
\midrule
GR00T-N1.7~\citep{bjorck2025gr00t}
& 98.20 & 99.00 & 98.00 & 88.40 & 95.90 \\
\rowcolor{experimentrow}
\textbf{Flow2Act-GR00T}
& 99.40 & 99.80 & 99.80 & 94.80
& \mbox{\textbf{98.45}\,{\footnotesize(+2.55)}} \\
\addlinespace[2pt]
$\pi_{0.5}$~\citep{intelligence2025pi_}
& 98.80 & 98.20 & 98.00 & 92.40 & 96.85 \\
\rowcolor{experimentrow}
\textbf{Flow2Act-$\pi_{0.5}$}
& 99.60 & 98.80 & 97.60 & 93.00
& \mbox{\textbf{97.25}\,{\footnotesize(+0.40)}} \\
\addlinespace[2pt]
Xiaomi-Robotics-0~\citep{cai2026xiaomi}
& 98.80 & 100.00 & 98.80 & 97.20 & 98.70 \\
\rowcolor{experimentrow}
\textbf{Flow2Act-Xiaomi}
& 99.60 & 100.00 & 99.00 & 97.80
& \mbox{\textbf{99.10}\,{\footnotesize(+0.40)}} \\
\bottomrule
\end{tabularx}
\end{table}

\begin{table}[t]
\centering
\small
\renewcommand{\arraystretch}{1.12}
\setlength{\tabcolsep}{2.2pt}
\vspace{3pt}
\caption{LIBERO-Plus~\citep{fei2025libero} success rate ($\%$, $\uparrow$).}
\label{tab:exp-libero-plus}
\begin{tabularx}{\textwidth}{
    >{\raggedright\arraybackslash}p{0.34\textwidth}
    *{7}{>{\centering\arraybackslash}X}
    >{\centering\arraybackslash}p{0.14\textwidth}
}
\toprule
\textbf{Model}
& \textbf{Camera}
& \textbf{Robot}
& \textbf{Lang.}
& \textbf{Light}
& \textbf{Backgr.}
& \textbf{Noise}
& \textbf{Layout}
& \textbf{Total} \\
\midrule
X-VLA~\citep{zheng2026x}
& 23.40 & 89.70 & 75.70 & 88.20
& 96.00 & 62.70 & 71.80 & 71.40 \\
ABot-M0~\citep{yang2026abot}
& 60.40 & 67.90 & 86.40 & 96.20
& 91.60 & 86.40 & 82.60 & 80.50 \\
StarVLA-OFT~\citep{community2026starvla}
& 47.00 & 60.10 & 87.00 & 96.30
& 95.30 & 73.10 & 79.20 & 75.00 \\
\midrule
GR00T-N1.7~\citep{bjorck2025gr00t}
& 47.65 & 30.06 & 81.26 & 92.99
& 94.80 & 76.08 & 70.43 & 68.31 \\
\rowcolor{experimentrow}
\textbf{Flow2Act-GR00T}
& 60.03 & 80.06 & 86.99 & 97.64
& 96.47 & 78.14 & 80.85
& \mbox{\textbf{80.39}\,{\footnotesize(+12.08)}} \\
\addlinespace[2pt]
$\pi_{0.5}$~\citep{intelligence2025pi_}
& 78.40 & 73.60 & 80.80 & 96.20
& 94.10 & 89.00 & 84.50 & 84.40 \\
\rowcolor{experimentrow}
\textbf{Flow2Act-$\pi_{0.5}$}
& 84.60 & 92.32 & 87.57 & 95.97
& 93.87 & 90.38 & 86.30
& \mbox{\textbf{87.48}\,{\footnotesize(+3.08)}} \\
\addlinespace[2pt]
Xiaomi-Robotics-0~\citep{cai2026xiaomi}
& 40.22 & 55.63 & 89.02 & 94.51
& 90.43 & 86.78 & 75.87 & 74.50 \\
\rowcolor{experimentrow}
\textbf{Flow2Act-Xiaomi}
& 69.86 & 85.94 & 87.44 & 97.29
& 96.38 & 93.63 & 86.36
& \mbox{\textbf{87.85}\,{\footnotesize(+13.35)}} \\
\bottomrule
\end{tabularx}
\end{table}

\subsection{Main Results}
\label{sec:exp-results}

\noindent\textbf{LIBERO.}\quad
Tab.~\ref{tab:exp-libero} shows that \our improves the average success rate of all three backbones on LIBERO. GR00T-N1.7 gains 2.6 points (95.9$\to$98.5), $\pi_{0.5}$ gains 0.4 (96.9$\to$97.3), and Xiaomi-Robotics-0 gains 0.4 (98.7$\to$99.1). The suite-level results show how these aggregate improvements are distributed across tasks. For GR00T-N1.7, success increases on all four suites, with the largest gain on Long: 88.40\% to 94.80\%, an improvement of 6.40 points. Spatial, Object, and Goal improve by 1.20, 0.80, and 1.80 points, respectively. The gain therefore extends across task categories while being particularly large on the long-horizon suite.

For $\pi_{0.5}$ and Xiaomi-Robotics-0, the baseline averages already exceed 96\%, leaving less room for absolute improvement. Both variants gain 0.80 points on Spatial and 0.60 points on Long, and the Xiaomi variant preserves 100.00\% success on Object. The improvements are not uniform across every suite: $\pi_{0.5}$ decreases from 98.00\% to 97.60\% on Goal. Thus, the results support an aggregate benefit across backbones, together with some variation across individual task suites.

\noindent\textbf{LIBERO-Plus.}\quad
Tab.~\ref{tab:exp-libero-plus} evaluates the same LIBERO-trained policies under seven perturbation dimensions, without additional fine-tuning. The gains over the corresponding backbones are larger than on LIBERO: 12.1 points for GR00T-N1.7 (68.3$\to$80.4), 3.1 for $\pi_{0.5}$ (84.4$\to$87.5), and 13.4 for Xiaomi-Robotics-0 (74.5$\to$87.9). Robot initial-state perturbations produce the largest gains for all three backbones, at 50.00, 18.72, and 30.31 points, respectively. Camera perturbations also improve across all three, by 12.38, 6.20, and 29.64 points. These results indicate that the aggregate robustness gains include substantial improvements when the robot's initial state or viewing perspective changes.

GR00T-N1.7 improves across all seven perturbation dimensions. The other backbones show a more mixed pattern: $\pi_{0.5}$ gains 6.77 points under language perturbations, while Xiaomi-Robotics-0 gains 6.85 points under sensor noise. Small regressions remain, including decreases of 0.23 points in both Lighting and Background for $\pi_{0.5}$, and 1.58 points in Language for Xiaomi-Robotics-0. The higher Total scores therefore coexist with local decreases. Overall, the larger gains under distribution shifts support the practical value of the complete framework for robust manipulation; the component ablations below examine the contributions of geometry, motion supervision, and history.

\begin{table}[t]
\centering
\small
\renewcommand{\arraystretch}{1.2}
\setlength{\tabcolsep}{8pt}
\vspace{2pt}
\caption{RoboTwin 2.0~\citep{chen2025robotwin} success rate.}
\label{tab:exp-robotwin}
\begin{tabular}{lcc}
\toprule
\textbf{Model} & \textbf{Clean} & \textbf{Random} \\
\midrule
X-VLA~\citep{zheng2026x} & 72.8 & 72.8 \\
ABot-M0~\citep{yang2026abot} & 86.1 & 85.1 \\
StarVLA~\citep{community2026starvla} & 88.2 & 88.3 \\
RynnVLA-Latent~\citep{zhao2026rynnvla} & 88.7 & 88.4 \\
Galaxea G0.5~\citep{liu2026g0} & 93.7 & 92.8 \\
\midrule
$\pi_{0.5}$~\citep{intelligence2025pi_} & 82.7 & 76.8 \\
\rowcolor{experimentrow}
\textbf{\our-$\pi_{0.5}$} & \textbf{87.3}\,{\footnotesize(+4.6)} & \textbf{82.6}\,{\footnotesize(+5.8)} \\
\addlinespace[2pt]
LingBot-VLA 2.0~\citep{wu2026foundation} & 93.5 & 92.8 \\
\rowcolor{experimentrow}
\textbf{\our-LingBot} & \textbf{94.0}\,{\footnotesize(+0.5)} & \textbf{93.6}\,{\footnotesize(+0.8)} \\
\bottomrule
\end{tabular}
\end{table}

\begin{table}[t]
\centering
\small
\renewcommand{\arraystretch}{1.15}
\setlength{\tabcolsep}{4pt}
\caption{Real-world success rates across six tasks (20 trials each).}
\label{tab:exp-real-world}
\begin{tabularx}{\columnwidth}{
    l
    *{6}{>{\centering\arraybackslash}X}
}
\toprule
\textbf{Method} & \textbf{Fruit} & \textbf{Block} & \textbf{Stack} & \textbf{Object} & \textbf{Drawer} & \textbf{Shirt} \\
\midrule
GR00T-N1.7~\citep{bjorck2025gr00t} & 40 & 50 & 5 & 10 & 15 & 5 \\
$\pi_0$~\citep{black2024pi_0} & 60 & 60 & 25 & 35 & 30 & 25 \\
$\pi_{0.5}$~\citep{intelligence2025pi_} & 85 & 75 & 40 & 50 & 55 & 40 \\
\rowcolor{experimentrow}
\textbf{\our} & \textbf{100} & \textbf{95} & \textbf{80} & \textbf{90} & \textbf{80} & \textbf{70} \\
\bottomrule
\end{tabularx}
\end{table}

\noindent\textbf{RoboTwin 2.0.}\quad
Tab.~\ref{tab:exp-robotwin} extends the evaluation to bimanual manipulation under Clean and Randomized conditions. With $\pi_{0.5}$, \our raises Clean success from 82.7\% to 87.3\% and Randomized success from 76.8\% to 82.6\%, improving the mean from 79.8\% to 85.0\%. The gain is larger under Randomized conditions than under Clean conditions (5.8 versus 4.6 points), consistent with the robustness improvements observed on LIBERO-Plus.

The gains also extend to LingBot-VLA 2.0, whose baseline already achieves 93.5\% and 92.8\% success in the two conditions. \our increases these scores to 94.0\% and 93.6\%, respectively, with the mean rising from 93.2\% to 93.8\%. The absolute improvements are smaller than for $\pi_{0.5}$, but both backbones improve under both conditions. Together, these comparisons show that the observed benefits extend beyond single-arm tasks to the bimanual setting while retaining each backbone's native action formulation and generator.

\noindent\textbf{Real-world manipulation.}\quad
As shown in Tab.~\ref{tab:exp-real-world}, \our outperforms all baselines across all six real-world tasks. Relative to the strongest baseline, $\pi_{0.5}$, the largest gains occur on Stack Blocks and Object Transfer, where success increases from 40\% to 80\% and from 50\% to 90\%, respectively. These tasks test instructed stacking configurations and coordinated object-to-target transfers. Improvements are also observed on the other task categories: \our succeeds in all 20 Fruit Placement trials, achieves 95\% success on Block-to-Cup, and reaches 80\% on Drawer Storage. These outcomes use the complete-task criterion, which requires correct object selection, execution order, and completion of all instructed stages rather than partial progress.

\begin{figure}[t]
\centering
\includegraphics[width=\linewidth]{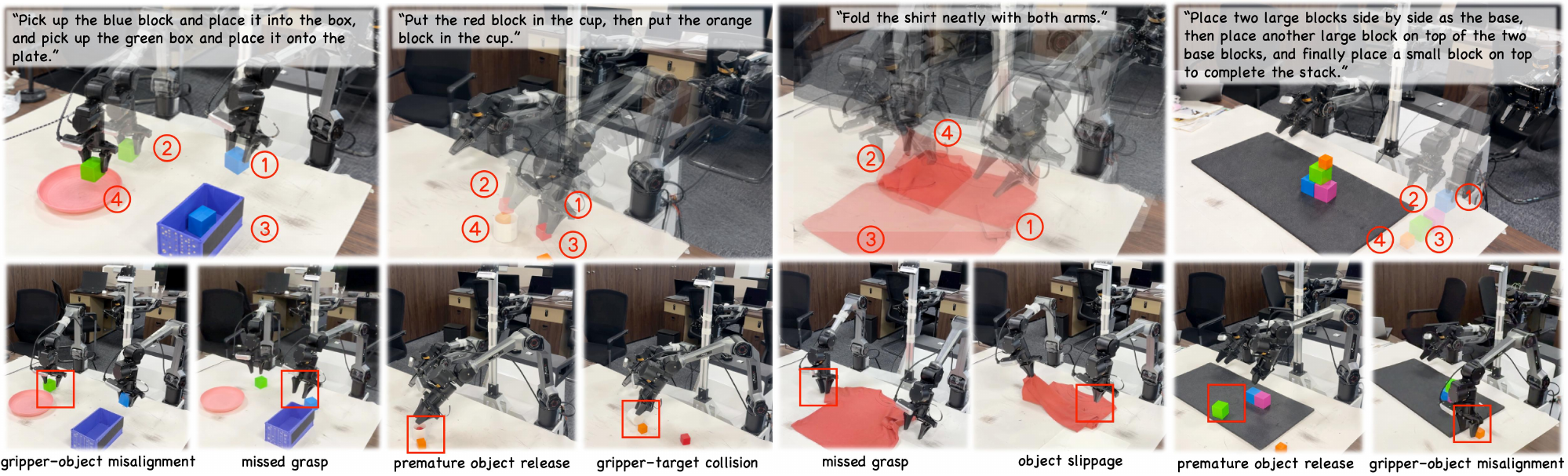}
\vspace{1pt}
\caption{\textbf{Qualitative results on real-world manipulation tasks.}
Top: overlaid frames illustrate task execution, with numbered markers indicating key stages. Bottom: representative failure cases from other models during deployment, including gripper--object misalignment, missed grasps, premature release, gripper--target collisions, and object slippage. Red boxes highlight failure regions.}
\label{fig:exp-real-data}
\end{figure}

Shirt Folding remains the most challenging task for \our, with 70\% success compared with 40\% for $\pi_{0.5}$. Its lower success rate leaves room for improvement in coordinated, multi-stage bimanual execution. Fig.~\ref{fig:exp-real-data} complements the task-level results with representative executions and baseline failure cases, including gripper--object misalignment, premature release, and missed grasps. The execution sequences in Appendix~\ref{app:real-world-executions} provide further examples of the instructed tasks. Taken across the six categories, the results show improvements on tasks involving object selection, ordered placement, storage, and bimanual coordination.

\subsection{Ablation Studies}
\label{sec:exp-ablation}

\subsubsection{Module Contributions}
\label{sec:exp-module-ablation}
We conduct ablations with GR00T-N1.7~\citep{bjorck2025gr00t} and $\pi_{0.5}$~\citep{intelligence2025pi_} on LIBERO and LIBERO-Plus.
Tab.~\ref{tab:exp-ablation} examines module contributions, motion supervision modalities, aligned layers, memory length, and adapter design.
Four aligned layers, $L=3$, and a shared convolutional adapter yield the best overall performance among the tested settings. The following analyses examine these design choices in detail.

\noindent\textbf{Module contributions and motion modalities.}
Adding geometry grounding, future flow supervision, and memory progressively improves both backbones.
On LIBERO-Plus, geometry grounding yields gains of 5.1 and 1.1 points for GR00T-N1.7 and $\pi_{0.5}$, respectively.
The additional gains from future flow suggest that explicit motion supervision benefits policy learning beyond static geometry grounding alone.
Future flow supervision adds another 4.5 and 0.9 points, followed by gains of 2.5 and 1.1 points from memory.
In the motion-modality ablation, paired 2D/3D supervision outperforms either modality alone on LIBERO-Plus, improving success by 3.9 and 0.8 points over the better single-modality variant for the two backbones, respectively.
These results suggest that image-plane trajectories and 3D displacements provide complementary motion supervision.

\begin{table}[!t]
\centering
\small
\setlength{\tabcolsep}{6pt}
\caption{\textbf{Ablation studies} on LIBERO~\citep{liu2023libero} and LIBERO-Plus~\citep{fei2025libero}. Each group varies one factor while fixing the rest at the reference configuration.}
\vspace{2pt}
\label{tab:exp-ablation}
\begin{tabular}{lcccc}
\toprule
& \multicolumn{2}{c}{\textbf{GR00T-N1.7}}
& \multicolumn{2}{c}{$\boldsymbol{\pi_{0.5}}$} \\
\cmidrule(lr){2-3}\cmidrule(lr){4-5}
\textbf{Configuration}
& LIBERO & LIBERO-Plus & LIBERO & LIBERO-Plus \\
\midrule

\multicolumn{5}{l}{\textit{Module ablation}} \\
\quad Baseline & 95.9 & 68.3 & 96.9 & 84.4 \\
\quad + Geometry Grounding & 97.3 & 73.4 & 97.1 & 85.5 \\
\quad + Future Flow & 98.1 & 77.9 & 97.2 & 86.4 \\
\rowcolor{experimentrow}
\quad + Memory
& \textbf{98.5} & \textbf{80.4} & \textbf{97.3} & \textbf{87.5} \\
\midrule

\multicolumn{5}{l}{\textit{Motion modality}} \\
\quad 2D only & 97.7 & 76.5 & 97.2 & 86.7 \\
\quad 3D only & 97.9 & 76.2 & 97.3 & 86.5 \\
\rowcolor{experimentrow}
\quad 2D + 3D
& \textbf{98.5} & \textbf{80.4} & \textbf{97.3} & \textbf{87.5} \\
\midrule

\multicolumn{5}{l}{\textit{Aligned layers}} \\
\quad 1 & 96.6 & 74.1 & 97.0 & 85.2 \\
\quad 2 & 97.4 & 75.7 & 97.1 & 85.6 \\
\rowcolor{experimentrow}
\quad 4
& \textbf{98.5} & \textbf{80.4} & \textbf{97.3} & \textbf{87.5} \\
\midrule

\multicolumn{5}{l}{\textit{Memory length $L$}} \\
\quad 1 & 98.1 & 77.9 & 97.2 & 86.4 \\
\quad 2 & 98.3 & 79.1 & 97.3 & 86.9 \\
\rowcolor{experimentrow}
\quad 3
& \textbf{98.5} & \textbf{80.4} & \textbf{97.3} & \textbf{87.5} \\
\quad 4 & 98.2 & 78.4 & 97.2 & 87.1 \\
\midrule

\multicolumn{5}{l}{\textit{Adapter design}} \\
\quad LN+MLP, Independent & 97.7 & 72.8 & 93.6 & 79.3 \\
\quad LN+MLP, Shared & 98.2 & 78.4 & 96.7 & 84.6 \\
\quad Conv, Independent & 97.9 & 76.5 & 95.3 & 84.1 \\
\rowcolor{experimentrow}
\quad Conv, Shared
& \textbf{98.5} & \textbf{80.4} & \textbf{97.3} & \textbf{87.5} \\

\bottomrule
\end{tabular}
\end{table}

\subsubsection{Design Choices}
\label{sec:exp-design-choices}

\noindent\textbf{Number of aligned layers.}
Increasing the number of aligned layers from one to two to four
changes GR00T-N1.7's LIBERO-Plus score from 74.1 to 75.7 to 80.4;
for $\pi_{0.5}$, the scores are 85.2, 85.6, and 87.5.
LIBERO scores also increase across these settings.
Four-layer alignment is consequently the best of the tested
choices for both backbones.
The experiment compares the extent of intermediate supervision;
it does not establish that adding layers beyond four would
continue to improve performance.

\noindent\textbf{Observation-context length.}
Here $L$ includes the current observation, so $L=1$ removes
historical observations while retaining future motion prediction.
GR00T-N1.7's LIBERO-Plus score increases from 77.9 at $L=1$
to 79.1 at $L=2$ and 80.4 at $L=3$, then falls to 78.4 at $L=4$.
For $\pi_{0.5}$, it changes from 86.4 to 86.9 to 87.5 to 87.1.
The results favor two past decision observations in the tested
range; a longer context is not uniformly better.
They do not distinguish whether the decline at $L=4$ is caused
by stale observations, optimization, or run-to-run variation.
The latency measurements in Tab.~\ref{tab:latency} additionally
show that re-encoding history increases runtime, making context
length relevant to both accuracy and deployment cost.

\noindent\textbf{Adapter architecture and parameter sharing.}
Sharing improves LIBERO-Plus performance for both adapter
architectures and both backbones.
For GR00T-N1.7, LN+MLP improves from 72.8 with independent
adapters to 78.4 with sharing, while Conv improves from 76.5
to 80.4.
For $\pi_{0.5}$, the corresponding changes are 79.3 to 84.6
and 84.1 to 87.5.
Among shared adapters, Conv exceeds LN+MLP by 2.0 and 2.9
points, respectively.
The shared convolutional adapter performs best among the four
tested combinations.
These comparisons establish the observed ranking but do not
independently identify whether parameter count, spatial
structure, or optimization explains it.

\subsubsection{History Sampling Strategy}
\label{sec:exp-history-sampling}
We fix $L=3$, comprising the current observation and two historical observations, and compare two training-time sampling strategies.
Decision-interval sampling selects observations at $t-E$ and $t-2E$, whereas random sampling uniformly selects two distinct timestamps from $[t-2E,t-1]$ within the same episode.
The selected historical features are jointly read with the current observation by the Future Motion queries.
Both variants use decision-interval sampling during evaluation, with all remaining training and evaluation settings held fixed under the same evaluation protocol.
Tab.~\ref{tab:historical-sample} compares the resulting policy performance.

\begin{table}[t]
\centering
\small
\setlength{\tabcolsep}{5pt}
\renewcommand{\arraystretch}{1.15}
\vspace{2pt}
\caption{Training-time history sampling with $L=3$. LIBERO and Plus report LIBERO Avg. and LIBERO-Plus Total, respectively.}
\label{tab:historical-sample}
\begin{tabularx}{\columnwidth}{
    l
    *{4}{>{\centering\arraybackslash}X}
}
\toprule
& \multicolumn{2}{c}{\textbf{GR00T-N1.7}}
& \multicolumn{2}{c}{$\boldsymbol{\pi_{0.5}}$} \\
\cmidrule(lr){2-3}\cmidrule(lr){4-5}
\textbf{Configuration} & LIBERO & Plus & LIBERO & Plus \\
\midrule
Random & 96.1 & 75.7 & 96.3 & 80.9 \\
\rowcolor{experimentrow}
\textbf{Decision-interval} & \textbf{98.5} & \textbf{80.4} & \textbf{97.3} & \textbf{87.5} \\
\bottomrule
\end{tabularx}
\end{table}

\subsubsection{Geometry Supervision Paradigm}
\label{sec:exp-geometry-query}
We compare direct alignment of native VLM visual features with \textit{Geometry Queries}, a variant that uses learnable queries to extract geometric representations.
In this variant, features from four intermediate-to-late VLM layers are projected and upsampled for a frozen DA3 depth decoder, and the predicted depth maps are supervised by
DA3 teacher outputs over valid pixels.
Our method instead aligns native VLM visual features with DA3 features through a shared adapter.
Both variants retain identical future-motion supervision, historical context, and action generation components.
As shown in Tab.~\ref{tab:geometry-interface}, direct alignment achieves higher success rates on LIBERO and LIBERO-Plus with both GR00T-N1.7 and $\pi_{0.5}$, yielding LIBERO-Plus
gains of 5.7 and 1.7 percentage points, respectively.
These results support our choice of direct feature alignment for geometric grounding.

\begin{figure}[t]
\centering
\begin{minipage}[t]{0.485\linewidth}
\centering
\includegraphics[width=\linewidth]{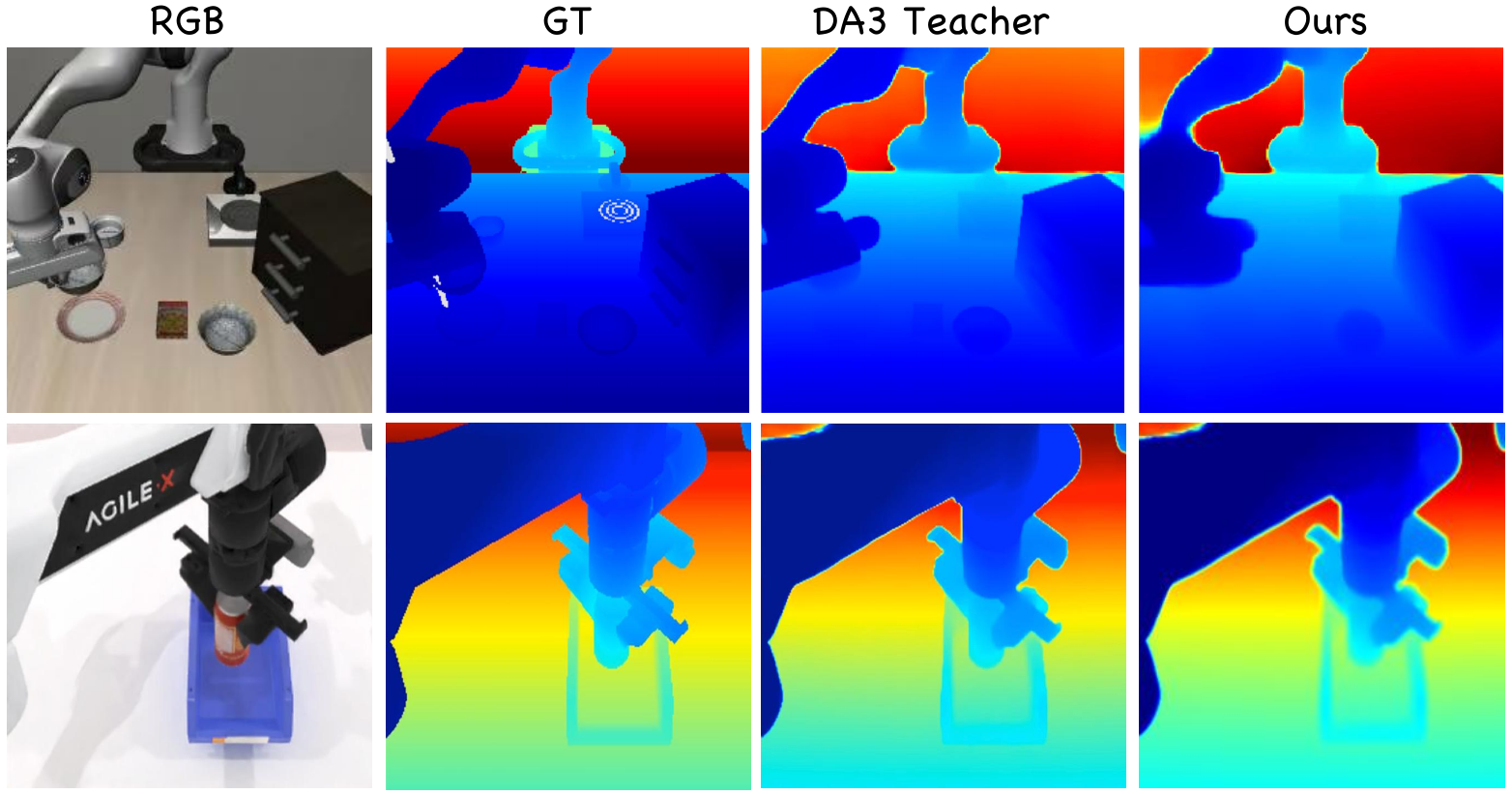}
\vspace{2pt}
\caption{\textbf{Depth comparisons} on LIBERO (top) and RoboTwin (bottom). From left to right: RGB, ground truth, DA3 teacher predictions, and ours.}
\label{fig:gt-teacher-stu-depth}
\end{minipage}\hfill
\begin{minipage}[t]{0.485\linewidth}
\centering
\includegraphics[width=\linewidth]{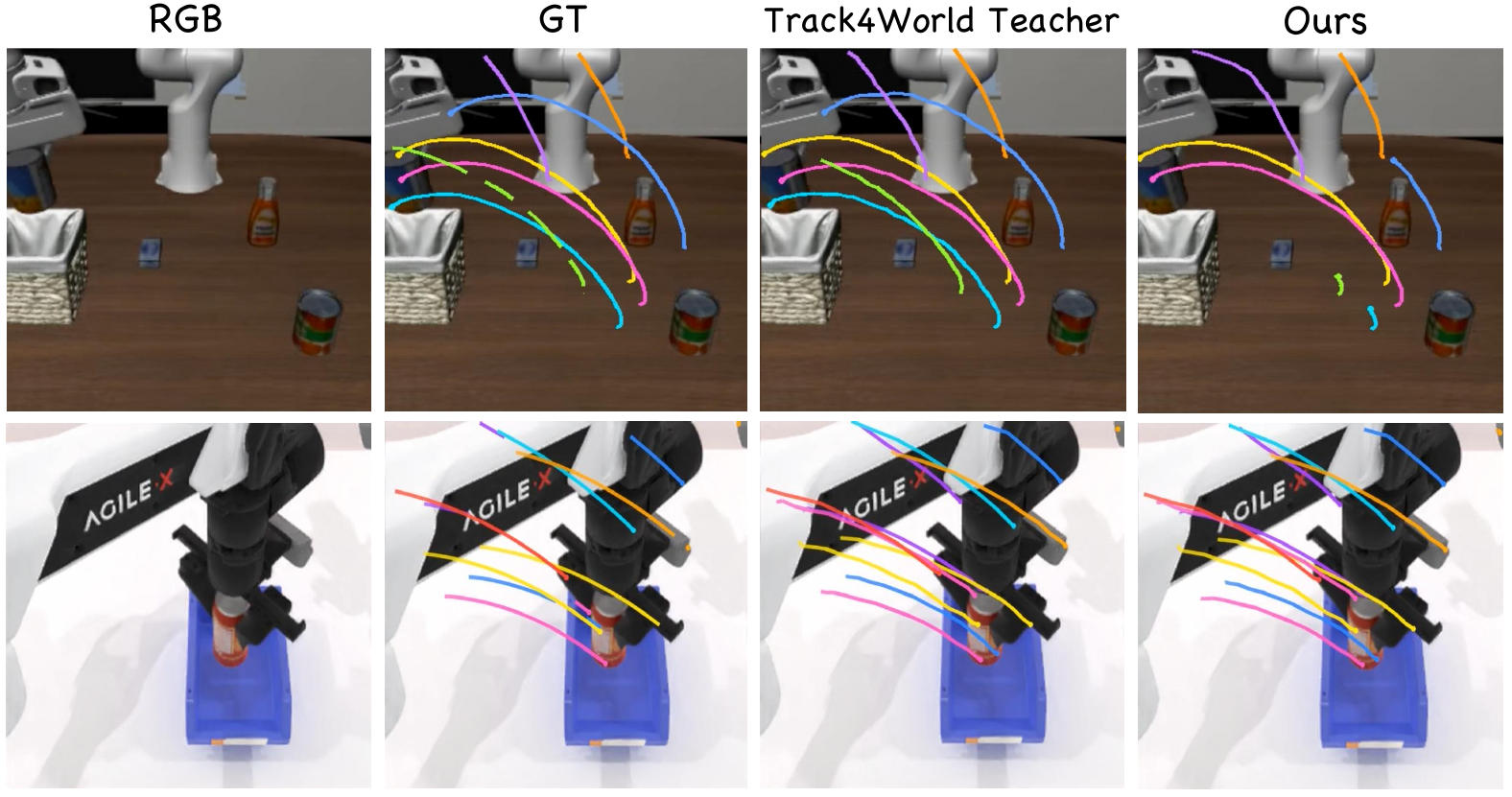}
\vspace{2pt}
\caption{\textbf{Future point-trajectory comparisons} on LIBERO (top) and RoboTwin (bottom). From left to right: RGB, ground truth, Track4World teacher predictions, and ours.}
\label{fig:gt-teacher-stu-motion}
\end{minipage}
\end{figure}

\begin{table}[t]
\centering
\small
\setlength{\tabcolsep}{5pt}
\renewcommand{\arraystretch}{1.15}
\vspace{2pt}
\caption{\textbf{Geometry supervision interface ablation} on LIBERO and LIBERO-Plus. We compare direct alignment of native VLM visual features with a query-based geometry supervision alternative.}
\label{tab:geometry-interface}
\begin{tabularx}{\columnwidth}{
    l
    *{4}{>{\centering\arraybackslash}X}
}
\toprule
& \multicolumn{2}{c}{\textbf{GR00T-N1.7}}
& \multicolumn{2}{c}{$\boldsymbol{\pi_{0.5}}$} \\
\cmidrule(lr){2-3}\cmidrule(lr){4-5}
\textbf{Configuration} & LIBERO & Plus & LIBERO & Plus \\
\midrule
Geometry Queries & 97.2 & 74.7 & 97.1 & 85.8 \\
\rowcolor{experimentrow}
\textbf{Direct Alignment} & \textbf{98.5} & \textbf{80.4} & \textbf{97.3} & \textbf{87.5} \\
\bottomrule
\end{tabularx}
\end{table}

\subsubsection{Effect of Geometry Grounding}
\label{sec:exp-geometry-grounding}
Geometry Grounding produces more distinct object boundaries and scene depth structure in the qualitative comparisons, complementing the policy-success ablations. Appendix~\ref{app:geometry-grounding-qualitative} provides the visual comparison with the DA3 pseudo-depth reference.

\subsection{Representation and Runtime Analysis}
\label{sec:exp-representation-runtime}

\subsubsection{Geometry and Motion Validation}

To assess teacher reliability and the quality of learned representations, we evaluate geometric estimations and motion predictions against simulator ground truth on LIBERO and RoboTwin.
As shown in Tab.~\ref{tab:geometry-motion-validation}, our method approaches the corresponding teachers' performance on both benchmarks.
Relative to the teachers, our depth RMSE is higher by 0.018\,m on LIBERO and 0.019\,m on RoboTwin, with corresponding 3D EPE gaps of 0.002\,m and 0.006\,m.
Notably, the student achieves this agreement without access to either teacher network during deployment.
Fig.~\ref{fig:gt-teacher-stu-depth} and~\ref{fig:gt-teacher-stu-motion} provide qualitative comparisons of depth and future point trajectories, respectively. Metric definitions and coordinate conventions are given in Sec.~\ref{app:geometry-motion-metrics}. Together, these results suggest that the student learns useful representations of both scene geometry and future point motion from RGB observations.

\noindent\textbf{Visualizations of 3D geometry and future motion flow.}
Fig.~\ref{fig:exp-depth-motion} visualizes depth and future motion predictions from RGB observations. The depth maps capture scene structure, while the trajectories
depict anticipated motion of current scene points over the action horizon.
These predictions illustrate joint geometric grounding and motion anticipation within the VLM at inference.

\begin{table}[t]
\centering
\small
\setlength{\tabcolsep}{4pt}
\renewcommand{\arraystretch}{1.15}
\vspace{2pt}
\caption{\textbf{Geometry and motion validation} against simulator ground truth. DA3 and Track4World provide depth and motion teacher targets, respectively.}
\label{tab:geometry-motion-validation}
\begin{tabularx}{\columnwidth}{
    @{}ll
    *{4}{>{\centering\arraybackslash}X}
    @{}
}
\toprule
& & \multicolumn{2}{c}{\textbf{Depth}}
  & \multicolumn{2}{c}{\textbf{3D Flow}} \\
\cmidrule(lr){3-4}\cmidrule(lr){5-6}
\textbf{Dataset} & \textbf{Method}
& \textbf{AbsRel} $\downarrow$
& \textbf{RMSE} $\downarrow$
& \textbf{EPE} $\downarrow$
& $\boldsymbol{\delta_{3D}^{0.05}}$ $\uparrow$ \\
\midrule
\multirow{2}{*}{LIBERO}
& Teacher & 0.241 & 0.327 & 0.080 & 51.7 \\
& Ours    & 0.252 & 0.345 & 0.082 & 48.3 \\
\midrule
\multirow{2}{*}{RoboTwin}
& Teacher & 0.154 & 0.112 & 0.106 & 47.9 \\
& Ours    & 0.183 & 0.131 & 0.112 & 45.2 \\
\bottomrule
\end{tabularx}
\end{table}

\begin{figure}[t]
\centering
\begin{minipage}[t]{0.49\linewidth}
\vspace{0pt}
\centering
\includegraphics[width=0.8\linewidth]{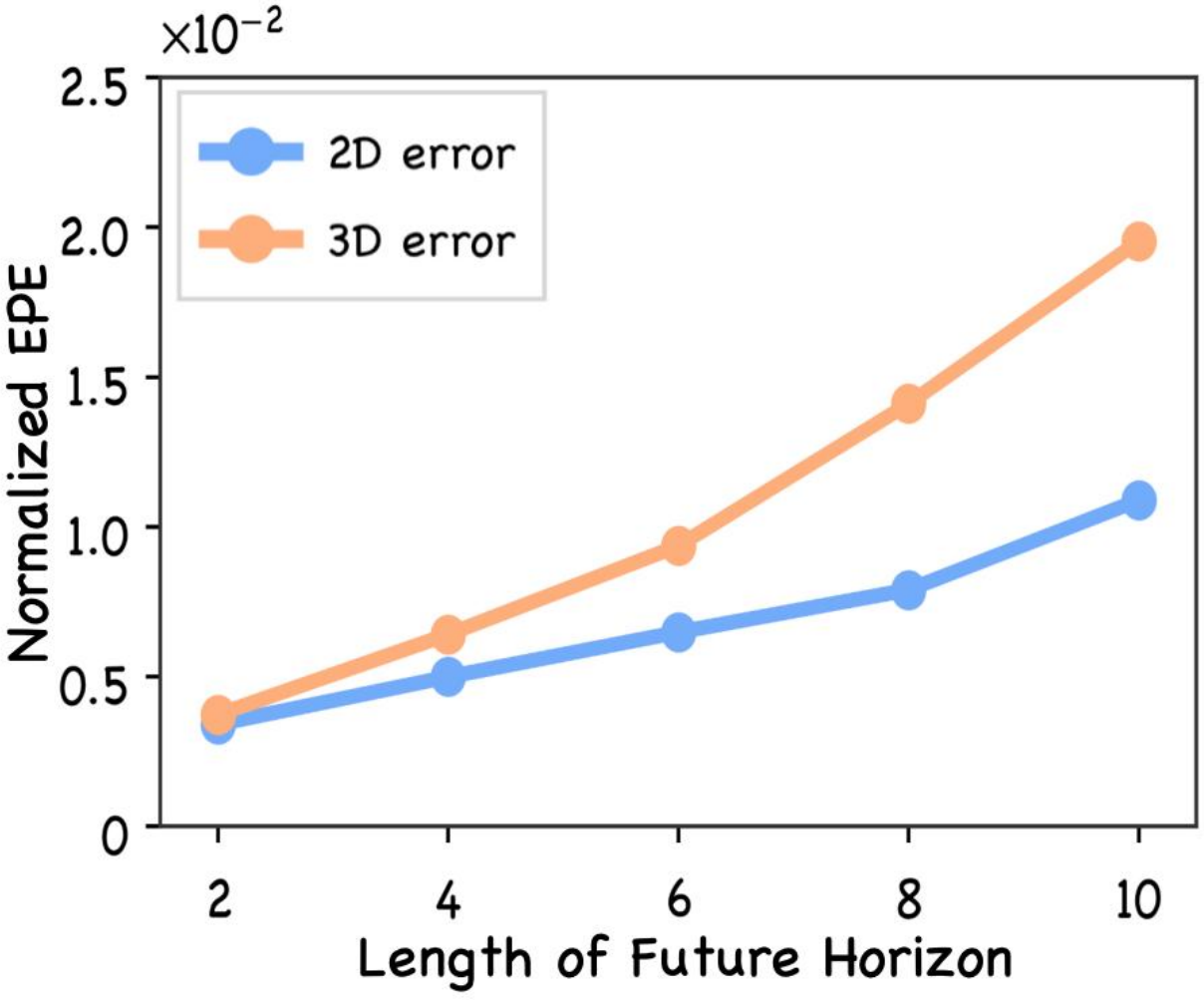}
\vspace{2pt}
\caption{\textbf{Motion prediction error across future horizons.}
Although both 2D and 3D endpoint errors increase with the prediction horizon, they remain low on their respective normalized scales.}
\label{fig:horizon-diagnostics}

\par\vspace{8pt}
\small
\setlength{\tabcolsep}{4pt}
\renewcommand{\arraystretch}{1.1}
\captionof{table}{\textbf{Inference latency} on a single NVIDIA A800 GPU. $L$ denotes the memory length, including the current observation.}
\label{tab:latency}
\begin{tabular}{lcc}
\toprule
\textbf{Configuration} & \textbf{Latency (ms)} & \textbf{Increase (ms)} \\
\midrule
Baseline & 206.00 & - \\
+ Static & 207.60 & +1.60 \\
+ Motion, $L=1$ & 212.69 & +6.69 \\
+ Motion, $L=2$ & 307.02 & +101.02 \\
+ Motion, $L=3$ & 353.37 & +147.37 \\
\bottomrule
\end{tabular}
\end{minipage}\hfill
\begin{minipage}[t]{0.48\linewidth}
\vspace{0pt}
\centering
\includegraphics[width=\linewidth]{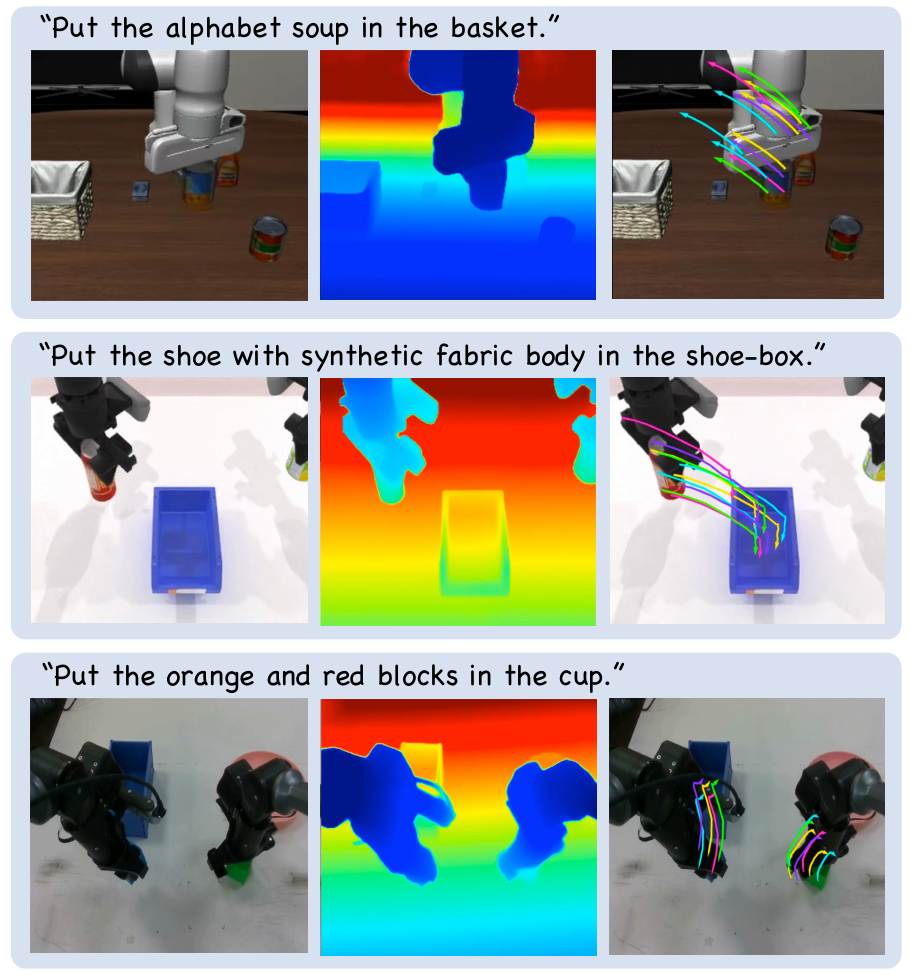}
\vspace{2pt}
\caption{\textbf{Qualitative geometry and motion predictions.} Rows show examples from LIBERO, RoboTwin 2.0, and real-world manipulation (top to bottom), with language instructions above each example. Columns show RGB observations, predicted depth maps, and
predicted future point trajectories (left to right).}
\label{fig:exp-depth-motion}
\end{minipage}
\end{figure}

\subsubsection{Horizon-Dependent Motion Prediction}
\label{sec:exp-diagnostics}
We evaluate how accurately the model predicts the motion of current scene points as the prediction horizon increases. On held-out episodes, we measure 2D and 3D normalized endpoint error (EPE) at future offsets $h\in\{2,4,6,8,10\}$. At each offset, we compare the predicted displacement of each valid anchor with its teacher-provided displacement from the same starting point. As shown in Fig.~\ref{fig:horizon-diagnostics}, both errors increase with the prediction horizon but remain small on their respective normalized scales: at $h=10$, the 2D and 3D EPE reach approximately 0.011 and 0.020, respectively. These results indicate that the predicted motion remains close to the teacher trajectories even at the longest evaluated horizon; they measure agreement with pseudo-labels rather than accuracy against simulator ground truth.

\subsubsection{Inference Latency}
We evaluate the runtime overhead of each component using $\pi_{0.5}$ as the baseline on a single NVIDIA A800 GPU. All configurations use BF16 precision, batch size 1, and a 10-step action horizon. As shown in Tab.~\ref{tab:latency}, geometry grounding introduces negligible overhead, as its teacher and adapter are inactive at inference. Future motion prediction adds little latency without historical context, whereas memory increases runtime due to re-encoding past observations.
All measurements include historical VLM encoding but exclude preprocessing, I/O, and communication. The 10-step runtime setting is distinct from the benchmark policy horizons in Tab.~\ref{tab:app-backbone-config}.

\section{Conclusion}
We presented \our, a framework that integrates geometric grounding and future motion anticipation into pretrained VLA policies. By combining future motion prediction with spatial geometric distillation, \our learns geometrically grounded future representations within the VLM. Sparse historical memory further provides temporal context, while future queries directly condition the native action generator through a residual attention bridge. Extensive evaluations demonstrate consistent improvements over the corresponding base policies, highlighting the benefits of integrating spatial and temporal reasoning into action generation.

\noindent\textbf{Limitations.}
Our approach relies on teacher-generated geometric and motion supervision, whose reliability may be affected by tracking errors and imperfect scene reconstruction. Moreover, fixed-interval historical sampling may not fully capture rapid or irregular interaction changes. Future work will explore more robust motion supervision and adaptive temporal memory to improve motion anticipation and long-horizon manipulation.

\clearpage
\bibliographystyle{assets/plainnat}
\bibliography{paper}

\clearpage
\beginappendix
\section{Additional Implementation Details}
\label{app:impl-details}

\subsection{Hyperparameters}
\label{app:hyperparams}
Table~\ref{tab:app-hyperparams} lists training settings accompanying
the benchmark policy configurations in Tab.~\ref{tab:app-backbone-config}.

\begin{center}
\begin{minipage}{\linewidth}
\centering
\small
\renewcommand{\arraystretch}{1.15}
\setlength{\tabcolsep}{6pt}
\captionof{table}{Shared training hyperparameters across all backbones and benchmarks.}
\label{tab:app-hyperparams}
\begin{tabular}{ll}
\toprule
\textbf{Hyperparameter} & \textbf{Value} \\
\midrule
Optimizer & AdamW ($\beta_1{=}0.9$, $\beta_2{=}0.95$) \\
LR (backbone) & $1\times10^{-5}$ \\
LR (adapters / heads) & $1\times10^{-4}$ \\
LR schedule & Cosine, 1000-step warmup \\
Batch size & 256 \\
Training steps & 40k (LIBERO), 80k (RoboTwin) \\
Weight decay & $1\times10^{-4}$ \\
$\lambda_{\mathrm{geo}}$ & 1.0 \\
$\lambda_{\mathrm{2D}}$ & 1.0 \\
$\lambda_{\mathrm{3D}}$ & 0.5 \\
$\mathcal L_{\mathrm{action}}$ coefficient & 1.0 \\
Gradient clipping & 1.0 \\
EMA decay & 0.999 \\
Precision & BF16 mixed \\
\bottomrule
\end{tabular}
\end{minipage}
\end{center}

\subsection{Supervision Details}
\label{app:supervision-construction}

\noindent\textbf{Paired supervision.}
Target construction follows Sec.~\ref{sec:future}. The 2D and 3D targets refer to the same current-image
anchor and future offset.
Future observations and teacher outputs are used only for
training supervision and are absent from deployment inputs.

\noindent\textbf{Validity and loss balancing.}
The 2D and 3D objectives have separate validity masks at each
anchor and future offset.
An unavailable 3D target does not discard an otherwise valid
2D correspondence.
The 2D objective balances the mean losses of moving and
low-motion anchors so that these groups enter the supervision
separately; the 3D objective averages over valid 3D targets.
An unavailable future target is distinct from an unavailable
historical observation: the former masks auxiliary supervision,
whereas the latter masks a context entry at episode startup.

\noindent\textbf{Feature distillation versus trajectory targets.}
Static supervision aligns intermediate DA3 features with student
visual features at the current camera views.
Motion supervision instead uses Track4World correspondences
and 3D displacements across time.
These are different teacher signals: the geometry adapter
serves the feature-alignment loss, while the two motion heads
serve displacement supervision.
Neither the teacher features nor the geometry adapter outputs
are supplied directly to the future queries or action generator.

\begin{figure}[t]
\centering
\includegraphics[width=0.65\linewidth]{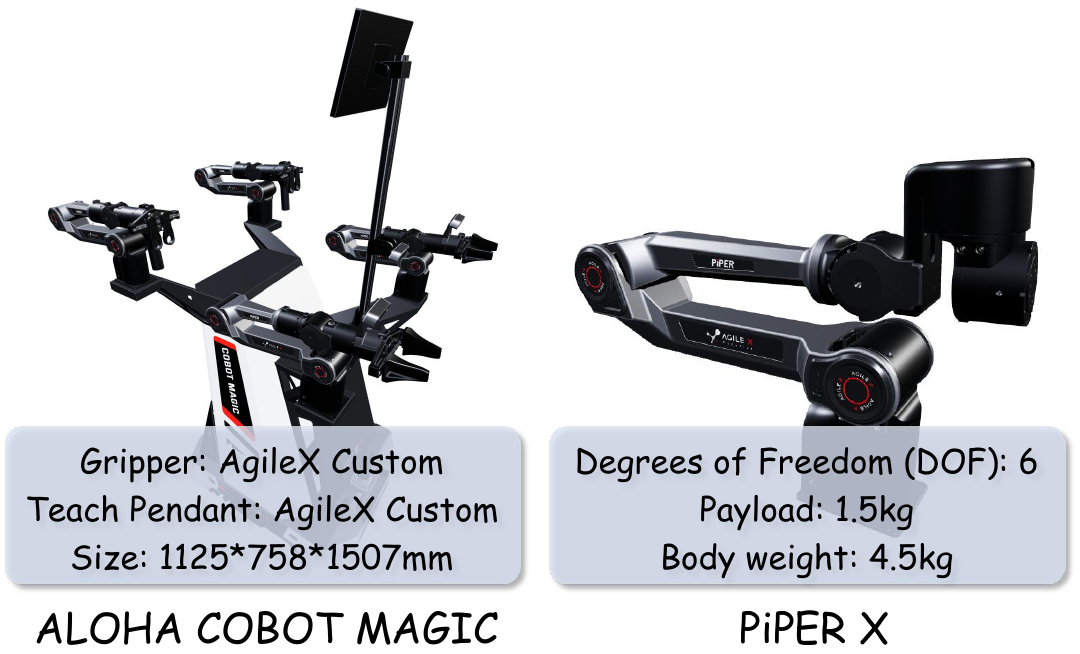}
\vspace{3pt}
\caption{Real-world robotic setup. The system consists of an ALOHA COBOT MAGIC data collection platform, a PiperX dual-arm robotic system, and three RGB cameras mounted on the head, left wrist, and right wrist, respectively.}
\label{fig:exp-real-robot}
\end{figure}

\subsection{Backbone Interfaces and Gradient Flow}
\label{app:backbone-interfaces}

\noindent\textbf{Backbone-dependent quantities.}
The action representation, native action loss, and action
generator are inherited from each pretrained backbone.
Tab.~\ref{tab:app-backbone-config} specifies the action horizon
$N$ and execution horizon $E$.
These quantities have different roles: $N$ determines the
number of future queries and the prediction horizon, whereas
$E$ determines replanning and historical observation spacing.
Motion supervision spans the action horizon without requiring
the auxiliary motion heads to produce robot commands.

\noindent\textbf{Shared auxiliary interfaces.}
Each future query has the VLM hidden width and represents one
future offset.
The motion readout combines a projected query state, the
current main-view feature sampled at the anchor, and a spatial
encoding.
LayerNorm, a linear layer, and GELU form the shared readout;
separate linear heads output a 2D displacement and a 3D
displacement.
Readout parameters are shared across anchors and offsets.
Static alignment uses four student--teacher layer pairs and a
convolutional adapter shared across selected layers and views.
Teacher features are downsampled to the corresponding student
token grid before the cosine-distance objective is computed.

\noindent\textbf{Historical context.}
The context length includes the current observation. The reference
setting uses the current observation and two preceding decision
observations, sampled at the deployment replanning interval.
Historical visual features are extracted using the shared VLM
with gradients stopped through that extraction pass.
The memory projection and the main VLM pass remain trainable.
Spatial, camera, and observation-age encodings distinguish the
historical visual tokens, which are combined with historical
robot-state tokens.
Unavailable early-episode entries are masked and the buffer
is cleared at episode boundaries.

\noindent\textbf{Attention and action conditioning.}
The main VLM sequence contains the current image--language
prefix, historical context, current robot state, and future queries.
Under the causal mask, query $h$ attends to the context $X_t$
and query positions $1,\ldots,h$, including its own position.
Current-prefix tokens do not attend to the subsequent history
or queries.
The native action generator is conditioned on layer-wise
key--value states of the current image--language prefix and
future queries.
Historical information therefore reaches action generation
through the history-conditioned queries.
The frozen teachers, geometry adapter, and dense motion
readouts are training or analysis components; action execution
retains the student backbone, historical context construction,
future queries, and native action generator.

\subsection{Real-World Robotic Setup}
\label{app:real-world-setup}

Our real-world experiments use an ALOHA COBOT MAGIC data collection
platform equipped with a PiperX dual-arm robotic system, as illustrated
in Fig.~\ref{fig:exp-real-robot}. Three RGB cameras provide visual
observations: a head-mounted camera captures the overall workspace,
while cameras mounted on the left and right wrists capture local
observations around the end effectors. These views provide complementary
information about the scene, manipulated objects, and robot--object
interactions. The policy receives all three RGB views together with
the language instruction and robot proprioceptive states.

\Needspace{0.78\textheight}
\begin{figure}[t]
\centering
\includegraphics[width=0.95\linewidth]{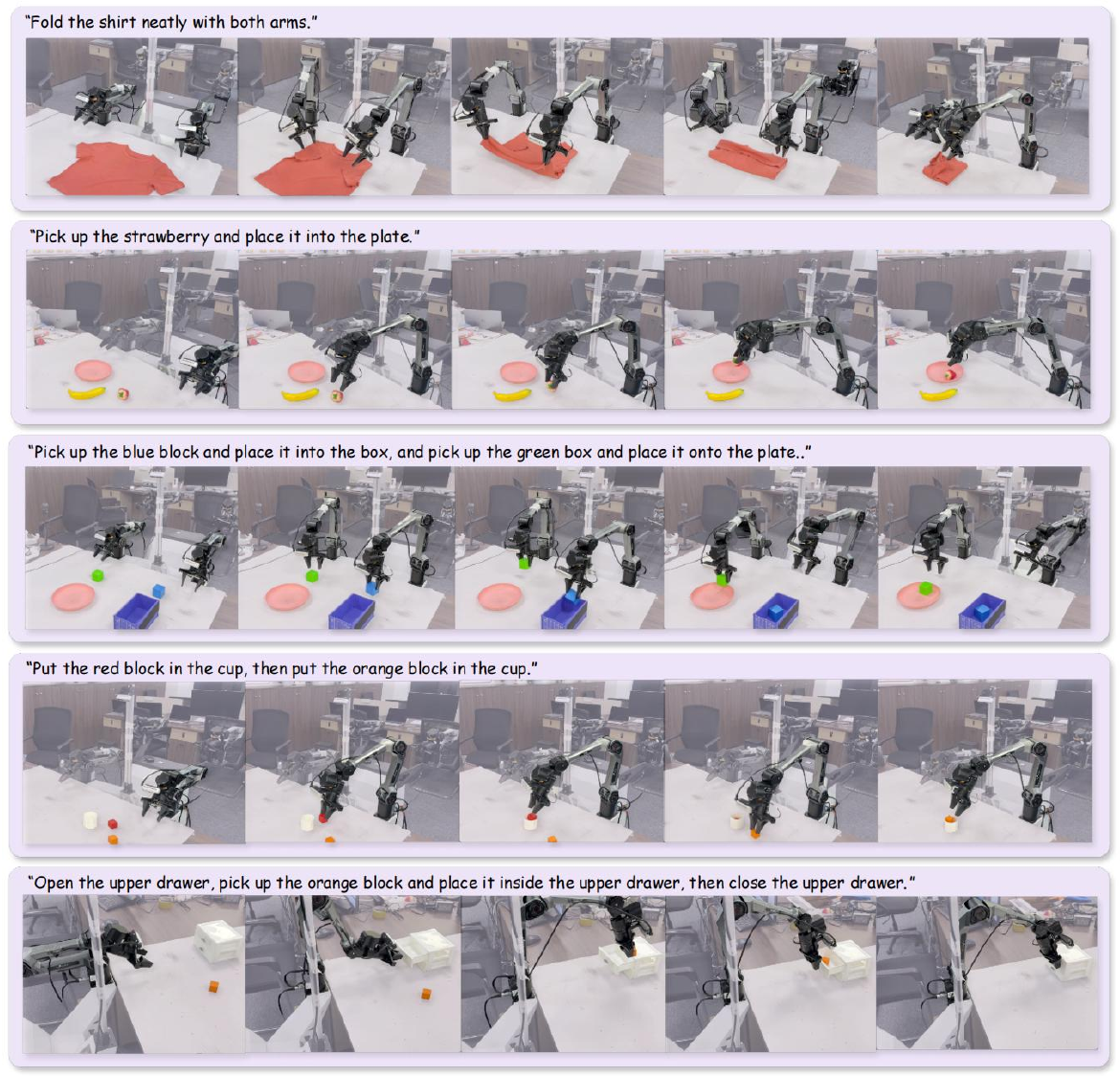}
\vspace{1pt}
\caption{Real-world manipulation sequences. From top to bottom: Shirt Folding, Fruit Placement, Object Transfer, Block-to-Cup, and Drawer Storage.}
\label{fig:exp-real-world}
\end{figure}

\subsection{Real-World Task Executions}
\label{app:real-world-executions}

Figure~\ref{fig:exp-real-world} presents representative execution
sequences for Shirt Folding, Fruit Placement, Object Transfer,
Block-to-Cup, and Drawer Storage. Each row follows the temporal
progression of a task under the language instruction shown above
the sequence, complementing the task descriptions in
Sec.~\ref{sec:exp-real-tasks}.

\section{Evaluation Protocol}
\label{app:eval-protocol}

\subsection{General Protocol}

All evaluations are closed-loop: at each decision step the policy receives the current observation and its decision history, predicts $N$ actions, and executes up to the first $E$ before replanning. Inference retains the student VLM, historical context construction, future queries, and native action generator; teacher networks, the geometry adapter, and dense auxiliary readouts are absent. The history bank is reset at every episode boundary.

\subsection{LIBERO}

We evaluate 50 episodes per task (500 episodes per suite, 2000 total). A trial is successful if the goal state is reached within the episode time limit. We report per-suite success rates and their unweighted average.

\subsection{LIBERO-Plus}

Policies trained exclusively on LIBERO are evaluated zero-shot. Each of the seven perturbation dimensions is tested independently; we report the success rate per dimension and the overall Total score as provided by the benchmark. Episode counts and task instances follow the official LIBERO-Plus release.

\subsection{RoboTwin 2.0}

We evaluate under Clean and Randomized conditions separately. A trial succeeds when the bimanual goal configuration is achieved within the time limit. We report Clean, Randomized, and their mean.

\subsection{Seeding and Reproducibility}

Our reported \our variants and ablation configurations each use a single training run and follow the evaluation protocol of the corresponding benchmark. Random seeds for environment resets follow the official benchmark splits. No post-hoc episode filtering or cherry-picking is applied.

\section{Geometry and Motion Evaluation}
\label{app:geometry-motion}

\subsection{Qualitative Effect of Geometry Grounding}
\label{app:geometry-grounding-qualitative}

Figure~\ref{fig:app-geometry-grounding-ablation} compares depth
visualizations with and without Geometry Grounding for two
scene observations.
Without grounding, the visualizations show blurred object
boundaries and less distinct scene structure.
With grounding, foreground contours and the spatial depth
variation more closely follow the DA3 reference, including
the robot silhouette and tabletop objects.
These examples qualitatively support the role of geometric
supervision in preserving scene structure, complementing the
policy-success comparisons above.

\begin{center}
\begin{minipage}{\linewidth}
\centering
\includegraphics[width=0.70\linewidth]{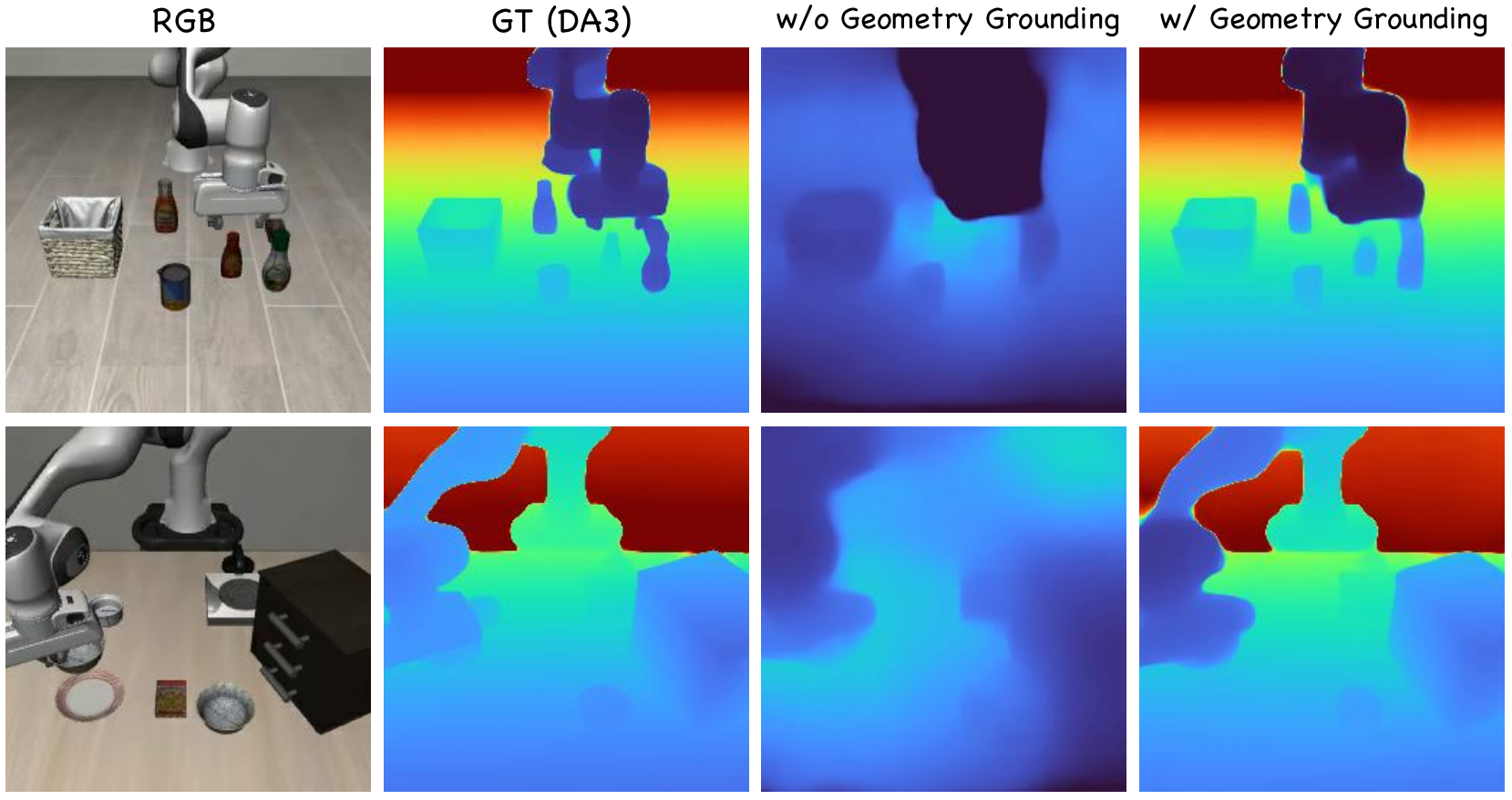}
\captionof{figure}{\textbf{Qualitative effect of Geometry Grounding.}
Columns show the RGB observation, DA3 reference depth, and
depth visualizations without and with Geometry Grounding,
respectively; rows show two scene examples.
Grounding produces more distinct foreground contours and
scene depth structure in these examples.
The figure label ``GT (DA3)'' denotes teacher-generated
pseudo-depth, not simulator ground truth.}
\label{fig:app-geometry-grounding-ablation}
\end{minipage}
\end{center}

\subsection{Depth and Motion Metrics}
\label{app:geometry-motion-metrics}

We evaluate depth and 3D motion against simulator ground truth
on LIBERO and RoboTwin, as reported in
Tab.~\ref{tab:geometry-motion-validation}.
DA3 supplies the depth-teacher predictions and
Track4World~\citep{lu2026track4world} supplies the motion-teacher predictions; neither teacher output
is used as the evaluation reference for this table.
Both teacher and student predictions are compared with the
corresponding simulator measurements.

\noindent\textbf{Depth evaluation.}
Let $\widehat D(p)$ be a teacher or student depth prediction and
$D^{\mathrm{gt}}(p)$ the simulator depth at the corresponding pixel.
Over valid pixels $\Omega_D$ with positive ground-truth depth,
we report
\begin{equation}
\begin{aligned}
 \mathrm{AbsRel}
 &=\frac{1}{|\Omega_D|}\sum_{p\in\Omega_D}
   \frac{|\widehat D(p)-D^{\mathrm{gt}}(p)|}{D^{\mathrm{gt}}(p)},\\
 \mathrm{RMSE}
 &=\left[\frac{1}{|\Omega_D|}\sum_{p\in\Omega_D}
   (\widehat D(p)-D^{\mathrm{gt}}(p))^2\right]^{1/2}.
\end{aligned}
\label{eq:app-depth-metrics}
\end{equation}
AbsRel is dimensionless; RMSE is reported in meters.
Lower values indicate better agreement with simulator depth.

\noindent\textbf{Ground-truth motion.}
For a point anchored at current-image pixel $p$, let
$P^{\mathrm{gt}}_{t,h}(p)$ denote its simulator position at time
$t+h$, expressed in a common evaluation coordinate frame.
The ground-truth displacement is
\begin{equation}
 \Delta P^{\mathrm{gt}}_{t,h}(p)
 =P^{\mathrm{gt}}_{t,h}(p)-P^{\mathrm{gt}}_{t,0}(p).
 \label{eq:app-gt-displacement}
\end{equation}
All offsets refer to the same current point and reference time.
The quantity being evaluated is cumulative displacement from
$t$ to $t+h$, rather than the displacement between consecutive
future frames.

\noindent\textbf{Prediction scale and coordinate convention.}
The student is trained to predict relative-scale displacement
$\widehat\delta_{t,h}(p)$.
Reversing the training normalization gives
$r_t\widehat\delta_{t,h}(p)$ in the Track4World reconstruction
scale, where $r_t$ is the median valid current-view depth from
that reconstruction.
This reversal is distinct from conversion to simulator metric
scale and coordinates.
In the metric definitions below, $\widehat{\Delta P}_{t,h}(p)$
denotes the teacher or student displacement expressed in the
same metric coordinate frame as $\Delta P^{\mathrm{gt}}_{t,h}(p)$.
A relative-scale output cannot be compared directly with a
meter-valued reference or a $0.05$\,m threshold.

\noindent\textbf{Motion metrics.}
For each valid anchor--offset pair, define
\begin{equation}
 e_{t,h}(p)=\left\|
 \widehat{\Delta P}_{t,h}(p)-\Delta P^{\mathrm{gt}}_{t,h}(p)
 \right\|_2 .
 \label{eq:app-point-error}
\end{equation}
Let $\Omega_F$ be the set of valid evaluated pairs.
The mean 3D endpoint error and threshold accuracy are
\begin{equation}
\begin{aligned}
 \mathrm{EPE}_{\mathrm{3D}}
 &=\frac{1}{|\Omega_F|}\sum_{(p,h)\in\Omega_F}e_{t,h}(p),\\
 \delta_{\mathrm{3D}}^{0.05}
 &=\frac{100}{|\Omega_F|}\sum_{(p,h)\in\Omega_F}
   \mathbf{1}\{e_{t,h}(p)<0.05\,\mathrm{m}\}.
\end{aligned}
\label{eq:app-flow-metrics}
\end{equation}
EPE is reported in meters and threshold accuracy as a percentage.
Lower EPE and higher threshold accuracy indicate better motion
prediction relative to simulator ground truth.
The teacher rows therefore measure teacher accuracy against
ground truth, rather than agreement of a teacher with itself.

\noindent\textbf{Horizon-dependent teacher agreement.}
The ground-truth validation above is separate from the
horizon-dependent representation diagnostic in
Fig.~\ref{fig:horizon-diagnostics}.
That curve measures normalized 2D/3D endpoint errors on
held-out episodes relative to teacher-provided trajectories at offsets
$h\in\{2,4,6,8,10\}$.
It assesses agreement with training pseudo-labels, whereas
Tab.~\ref{tab:geometry-motion-validation} assesses accuracy
against simulator ground truth.
The reference trajectories, normalization, and units of these
two evaluations should therefore be distinguished.
Depth and motion readouts are used for supervision and
analysis and are absent during action execution.

\end{document}